\documentclass{article}

\usepackage{arxiv}

\usepackage[utf8]{inputenc} 
\usepackage[T1]{fontenc}    
\usepackage{hyperref}       
\usepackage{url}            
\usepackage{booktabs}       
\usepackage{amsfonts}       
\usepackage{nicefrac}       
\usepackage{microtype}      

\usepackage{lipsum}
\usepackage{graphicx}
\usepackage{multirow}
\usepackage{color,array,dcolumn}
\usepackage{amsmath}
\usepackage{amssymb}
\usepackage[export]{adjustbox}

\newcolumntype{L}[1]{>{\raggedright\let\newline\\\arraybackslash\hspace{0pt}}m{#1}}
\newcolumntype{C}[1]{>{\centering\let\newline\\\arraybackslash\hspace{0pt}}m{#1}}
\newcolumntype{R}[1]{>{\raggedleft\let\newline\\\arraybackslash\hspace{0pt}}m{#1}}

\title{Joint learning of variational representations and solvers for inverse problems with partially-observed data}
\author{
  R. Fablet\thanks{Corresponding author)} \\
  IMT Atlantique, UMR CNRS Lab-STICC\\
  Brest, FR \\
  \texttt{ronan.fablet@imt-atlantique.fr} \\
   \And
  L. Drumetz \\
  IMT Atlantique, UMR CNRS Lab-STICC\\
  Brest, FR \\
  \texttt{lucas.drumetz@imt-atlantique.fr} \\
   \And
  F. Rousseau \\
  IMT Atlantique, UMR INSERM Latim\\
  Brest, FR \\
  \texttt{francois.rousseau@imt-atlantique.fr} \\
}

\begin{document}
\maketitle

\begin{abstract}
Designing appropriate variational regularization schemes is a crucial part of solving inverse problems, making them better-posed and guaranteeing that the solution of the associated optimization problem satisfies desirable properties. Recently, learning-based strategies have appeared to be very efficient for solving inverse problems, by learning direct inversion schemes or plug-and-play regularizers from available pairs of true states and observations. In this paper, we go a step further and design an end-to-end framework allowing to learn actual variational frameworks for inverse problems in such a supervised setting. The variational cost and the gradient-based solver are both stated 
as neural networks using automatic differentiation for the latter. 
We can jointly learn both components 
to minimize the data reconstruction error on the true states. This  leads to a data-driven discovery of variational models. We consider an application to inverse problems with incomplete datasets (image inpainting and multivariate time series interpolation). We experimentally illustrate that this framework can lead to a significant gain in terms of reconstruction performance, including w.r.t. the direct minimization of the variational formulation derived from the known generative model.


\end{abstract}


\section{Introduction}
\label{sec:intro}

Solving an inverse problem consists in computing an acceptable solution of a model that can explain measured observations. Inverse problems involve so-called forward or generative models describing the generation process of the observed data. Inverting a defined forward model is frequently an ill-posed problem. This leads to identifiability issues, which means that naive inversion schemes based on direct least squares inversion do not yield a neither good nor unique solution. 
The variational framework is a popular approach to solve inverse problems by defining an energy or cost whose minimization leads to an acceptable solution. This cost can be broken down into two terms: a data fidelity term related to a specific observation model and a regularization term characterizing the space of acceptable solutions. From a Bayesian point of view, one finds a trade-off between the data fidelity term and the regularization term specified independently from a particular observation configuration. The solution sought is then the Maximum a Posteriori (MAP) estimate of the state, given the observation data. In such approaches, the prior or regularization term is specified regardless of the observation configuration.

\begin{figure}[tb]
    \centering
    \includegraphics[width=13cm]{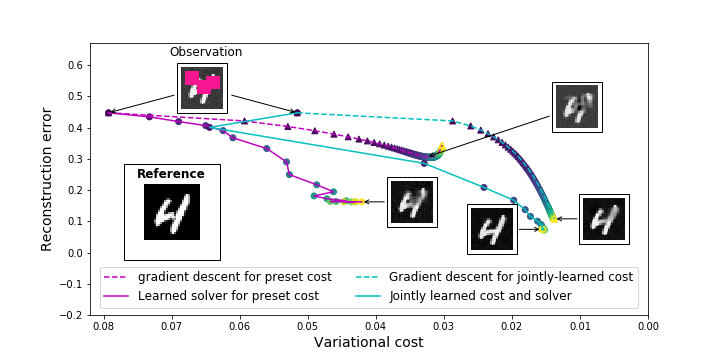}
    \caption{{\bf Solvers' energy pathways depending on the regularization term}:  when considering a predefined PCA-based regularization, the minimizer of the corresponding variational cost (magenta, dashed) may not converge to the expected hidden state especially for partially-observed data as illustrated here for an inpainting example. Using an end-to-end formulation, we may learn a gradient-based solver (magenta, solid) to improve the reconstruction performance. It may be further improved when 
    simultaneously learning the variational cost and the associated solver, here using a CNN-parameterized cost (cyan). In such cases, we expect a gradient descent (dashed) and the trained solver (solid) to converge to similar solutions, the latter in very few steps (here, 15). 
    }
    \label{fig: descent paths}
    \vspace*{-0.25cm}
\end{figure}

For a given inverse problem, it is then necessary to specify accurately the observation model, to define an appropriate regularization term and to implement a suitable optimization method. These three steps are the key elements in solving an inverse problem using variational approaches. An important issue is that there is no guarantees in general that the solution of the resulting optimization problem is actually the true state 
from which the observed data have been generated (see Fig.\ref{fig: descent paths}). This mismatch can have several causes: discrepancy between the forward model and the actual data generation process, unsuitable regularization term or optimization algorithm stuck in a local minimum.

In this article, we address the design and resolution of inverse problems as a meta-learning problem \cite{hospedales_meta-learning_2020}. 
With an emphasis on partial-observed data, our main contributions are as follows: (i) a versatile end-to-end framework for solving inverse problems by jointly learning the variational cost and the corresponding solver; (ii) experiments showing that learned iterative solvers can greatly speed up and improve the inversion performance compared to a gradient descent for predefined generative models; (iii) experiments showing that the joint learning of the variational representation and the associated solver may further improve the reconstruction performance, including when the true generative model is known. We also provide the accompanying Pytorch code for our results to be reproducible\footnote{The code of the preprint is available at \url{https://github.com/CIA-Oceanix/DinAE_4DVarNN_torch}}.

\section{Background}
\label{sec:pb}
The resolution of inverse problems is classically stated as the minimization of a variational cost \cite{aubert_mathematical_2006}
\begin{equation}
\displaystyle
\widehat{x} =\arg \min_{x\in {\cal{X}}} U_{O}(x,y) + U_{R}(x)
\end{equation}
where $x$ is the unknown state, $y$ an observation,  $U_{O}$ the data fidelity term which relates the unknown state to the observation and $U_{R}$ a regularization term. The latter aims at turning the ill-posed inversion of the observation into a solvable problem. 

As case-study, we focus on problems where the observation only conveys a partial information about the true state: especially $x$ and $y$ live in some space ${\cal{X}}$ and observation term writes as  $U_{O}(x,y)=\|x-y\|_\Omega^2$, where  $\Omega$ is the domain where observation $y$ is truely sampled\footnote{ $\|x\|_\Omega^2$ refer to the squared norm of $x$ over domain $\Omega$, e.g. $\|x\|_\Omega^2=\int_\Omega x(p)^2dp$ for scalar images.}.
$\Omega$ may for instance refer to a spatial domain in image interpolation and inpainting problems \cite{cheng_inpainting_2014}, to components of variable $x$ in matrix completion \cite{candes_2010} and data assimilation problems \cite{blum_data_2009,candes_2010}.  
Regarding the regularization term, a variety of approaches have been proposed, including gradient norms among which the total variation $U_{R}(x)=\|\nabla x\|$ \cite{aubert_mathematical_2006,alvarez_axioms_1993}, dictionary-based terms \cite{elad_image_2006,mairal_online_2009} or Markov Random Fields \cite{perez_markov_1998,geman_random_1990}. Most of these terms may be rewritten as $U_{R}(x)= \rho \left ( x -\Phi(x) \right )$ where $\rho$ is typically the $L_1$ or $L_2$ norm and $\Phi$ a projection-like operator. Overall, we consider a variational functional $U_\Phi$ for partially-observed data given by  
\begin{equation}
\label{eq: var model}
\displaystyle
U_\Phi \left ( x , y , \Omega \right ) = \lambda_1 \left \|x-y\right \|^2_\Omega + \lambda_2 \left \|x - \Phi(x) \right \|^2
\end{equation}
where $\lambda_{1,2}$ are preset or learnable scalar weights. 
Given some parameterization for operator $\Phi$, the minimization of functional $U_\Phi$ usually involves iterative gradient-based algorithms, for instance iterative updates given by $x^{(k+1)} = x^{(k)} - \alpha \nabla_x U_\Phi  ( x^{(k)} , y , \Omega )$
where $\nabla_x U_\Phi$ is the gradient operator of function $U_\Phi$ w.r.t. state $x$, typically derived using Euler-Lagrange equations \cite{aubert_mathematical_2006}.
For inverse problems with time-related processes, one may consider the adjoint method, at the core of variational data assimilation schemes in geoscience \cite{blum_data_2009}. Interestingly, using automatic differentiation tools as embedded in deep learning frameworks, the computation of this gradient operator is straightforward given that operator $\Phi$ can be stated as a composition of elementary operators such as tensorial and convolution operators and non-linear componentwise activation functions.

As detailed hereafter, this property provides us the basis for designing fully-learnable end-to-end formulations, which combine an explicit variational representation (\ref{eq: var model}) and an iterative gradient-based solver (\ref{eq: lstm update}) to address the reconstruction of the unknown state $x$ from observation $y$. 

\section{End-to-end learning framework}
\label{sec:learning}

This section presents the proposed end-to-end  framework for the joint learning of variational representations and associated solvers for inverse problems with partial observations. We first introduce in Section \ref{ss: operator phi} the considered parameterizations for operator $\Phi$ in (\ref{eq: var model}), the proposed end-to-end architecture in Section \ref{ss: EtoE} and the learning strategy in Section \ref{ss: learning}.

\subsection{Neural network representations for operator $\Phi$}
\label{ss: operator phi}

The design of neural network (NN) architectures for operator $\Phi$ is a key component of the proposed framework.
We may distinguish 
physics-informed and purely data-driven representations. Regarding the latter, dictionary-based or auto-encoder representations \cite{elad_image_2006} lead to operator $\Phi$ parameterized as $\Phi(x) = \phi ( \psi(x)  )$
with $\psi$ a mapping from the original space to a lower-dimensional space and $\phi$ the inverse mapping. 

Here, the assumption that the solution lies in a lower-dimensional space may be fairly restrictive and typically result in smoothing out fine-scale features in signals and images. We may rather exploit a 
NN architecture whose inputs and outputs have the same dimension as state $x$, including any architecture $\Psi$ proposed for directly solving the targeted inverse problem as $x=\Psi(y)$ as in plug-an-play approaches \cite{meinhardt2017learning,lunz2018adversarial,xie_image_2012}.
As an example, for the sake of simplicity, we consider across all case-studies a two-scale CNN architecture (subsequently denoted 2S-CNN): 
\begin{equation}
\label{eq: 2-scale GENN}
    \Phi(x) = Up \left ( \Phi_1 \left ( Dw(x) \right )\right ) + \Phi_2(x)
\end{equation}
where $Up$ and $Dw$ are upsampling (ConvTranspose layer) and downsampling (AveragePooling layer) operators. $\Phi_{1,2}$ are CNNs combining convolutional layers and ReLu activations.  

Regarding physics-informed representations, 
numerous recent studies have stressed the design of neural networks based on ordinary and partial differential equations (ODE/PDE) \cite{raissi2019physics,chen_neural_2018}. 
Especially, neural ODE representations come to parameterize operator $\Phi$ as $\Phi(x)(t+\Delta)=\phi ( x(t) ) $with $\Delta$ the integration time step and $\phi$ the one-step-ahead integration of the considered ODE.



\subsection{End-to-end architecture}
\label{ss: EtoE}

The proposed end-to-end architecture consists in embedding an iterative gradient-based solver based on the considered variational representation (\ref{eq: var model}). As inputs, we consider an observation $y$, the associated observation domain $\Omega$ and some initialization $x^{(0)}$. Following meta-learning schemes \cite{andrychowicz_learning_2016}, the gradient-based solver involves a residual architecture using a LSTM. More precisely, the $k^{th}$ iterative update within the iterative solver is given by
\begin{equation}
\label{eq: lstm update}
\left \{\begin{array}{ccl}
     g^{(k+1)}& = &  LSTM \left[ \alpha \cdot \nabla_x U_\Phi \left ( x^{(k)},y , \Omega\right),  h(k) , c(k) \right ]  \\~\\
     x^{(k+1)}& = & x^{(k)} - {\cal{H}}  \left( g^{(k+1)} \right )  \\
\end{array} \right.
\end{equation}
where $g^{(k+1)}$ is the output of an LSTM cell using as input gradient $\nabla_x U_\Phi ( x^{(k)},y , \Omega )$ for state estimate at iteration $k$,
$h(k),c(k)$ the internal states of the LSTM,
$\alpha$ a normalization scalar and ${\cal{H}}$ a linear or convolutional mapping. Let us denote by $\Gamma$ this iterative update operator and $\Psi_{\Phi,\Gamma}  (x^{(0)},y,\Omega )$ the output of the solver given initialization $x^{(0)}$, observation $y$ and observed domain $\Omega $.

Overall, the end-to-end architecture $\Psi_{\Phi,\Gamma}$ involves a predefined number of iterative updates typically from 5 to 20 in the reported experiments. We may consider other iterative solvers, including fixed-point procedures and other recurrent cells in place of the LSTM cells \cite{fablet_end--end_2019}. Experimentally, the latter was more efficient. The total number of parameters of the proposed end-to-end architecture comprises the parameters of operator $\Phi$, weights $\lambda_{1,2}$ and the parameters of gradient-based update operator $\Gamma$.

\subsection{Joint learning scheme}
\label{ss: learning}


Given pairs of observation data $(y_n,\Omega_n)$ and hidden states $x_n$, we state the joint learning of operator $\Phi$ and solver operator $\Gamma$ as the minimization of a reconstruction cost:
\begin{equation}
\label{eq: E2E loss}
   \arg \min_{\Phi,\Gamma} \sum_n {\cal{L}}\left (x_n,\tilde{x}_n \right) \mbox{  s.t.  } 
   \tilde{x}_n = \Psi_{\Phi,\Gamma}  (x_n^{(0)},y_n,\Omega _n)
\end{equation}
where ${\cal{L}} (x,\tilde{x}) = \nu_1 \|x-\tilde{x}\|^2 + \nu_2 \|x-\Phi(x)\|^2 + \nu_3 \|\tilde{x}-\Phi(\tilde{x})\|^2, \forall x,\forall y \in {\cal{X}}$ combines the reconstruction error of the true state with the projection errors of both the reconstructed state and the true state through weights $\nu_{1,2,3}$.
Given the  end-to-end architecture $\Psi_{\Phi,\Gamma}$, we can directly apply stochastic optimizers such as Adam to solve this minimization. 
Obviously, we may also consider a sequential learning scheme, where operator $\Phi$ is first set {\em a priori} or learned according to reconstruction loss
$\sum_n   \| x_n - \Phi  (x_n )  \|^2$, and the above minimization is carried out w.r.t. $\Gamma$ only, $\Phi$ being fixed.

All reported experiments have been run using Pytorch and the Adam optimizer for the learning stage. Regarding the gradient-based solver, we gradually increase the number of gradient-based iterations during the learning process, typically from 5 to 20. Weights $\nu_{1,2,3}$ are set to 1., 0.05 and 0.05. When jointly learning operators $\Phi$ and $\Gamma$, we use a halved learning rate for the parameters of operator $\Phi$.
 
\section{Results}
\label{sec:exp}

We run numerical experiments to illustrate and evaluate the proposed framework. We consider two different case-studies: image inpainting with MNIST data and the reconstruction of downsampled signals governed by ODEs. The latter provides us with means to investigate learnable variational formulations and solvers when the true model which governs the hidden states is known. 

\begin{table*}[tbh]
    \footnotesize
    \centering
    \begin{tabular}{|C{1.5cm}|C{1.5cm}||C{2cm}|C{1.75cm}|C{1.75cm}|C{1.75cm}|}
    \toprule
    \toprule
     \bf Model $\Phi$ &\bf Joint learning &\bf Solver&\bf R-Score&\bf I-score&\bf P-score\\
    \toprule
    \toprule
    DICT& No & OMP & 0.25 & 0.90 &  0.21 \\
     & No & Lasso & 0.20 & 0.71 & 0.21 \\
    \toprule
    \toprule
     & No & FSGD & 0.39 & 1.09 &  {\bf 0.12} \\
     PCA& No & LSTM-S & 0.15 & 0.54 & {\bf 0.12} \\
    \bottomrule
     & Yes & FP(1) & 0.4 & 1.51 & {\bf 0.12} \\
     AE& No & LSTM-S & 0.195 & 0.71 & {\bf 0.12} \\
     & Yes & LSTM-S & 0.210 & 0.76 & 0.95 \\
    \bottomrule
     2S-CNN& Yes & FP(1) & 0.16  & 0.56 & 0.03 \\
     & Yes & LSTM-S & {\bf 0.09} & {\bf 0.33} & 0.02 \\
    \bottomrule
    \bottomrule
    \end{tabular}
    \caption{{\bf MNIST experiment:} The training and test datasets comprise MNIST images corrupted by a Gaussian additive noise and three randomly-sampled 9x9 holes. We consider three parameterizations for operator $\Phi$ in the variational formulation defined by (\ref{eq: var model}),  a 50-dimensional PCA decomposition (PCA), a 20-dimensional dense auto-encoder (AE) and a bilinear 2S-CNN representation. We also compare a fixed-step gradient descent for variational energy (\ref{eq: var model}), referred to as FSGD, a one-step fixed-point iteration (FP(1)) and a LSTM-based solver (LSTM-S) based on the automatic differentiation of energy (\ref{eq: var model}). We may jointly train operator $\Phi$ and the LSTM solver or solely train the latter given a predefined operator $\Phi$. We also include two sparse coding schemes (DICT) adapted from \cite{mairal_online_2009}.
    As evaluation scores, we compute the reconstruction error for the entire image domain (R-score), the reconstruction error for missing data areas only (I-score), the projection error $x-\Phi(x)$ for gap-free and noise-free images (P-score). 
    We let the reader refer to the main text for additional information.}
    \label{tab:MNIST A}
\end{table*}

\subsection{MNIST data}

MNIST data provide a relatively simple image dataset 
to evaluate the proposed framework with different types of regularization terms in (\ref{eq: var model}). 
Especially, the MNIST dataset seems well-suited to dictionary and auto-encoder priors \cite{elad_image_2006,mairal_online_2009}.
We simulate observed images with 3 randomly sampled $9 \times 9$ missing data areas and an additive Gaussian noise of variance $0.1*\sigma^2$, with $\sigma^2$ the pixel-wise variance of the training data.
We consider three types of parameterizations for operator $\Phi$:
\begin{itemize}
    \item {\bf a linear PCA decomposition}: a 50-dimensional PCA operator which amounts to $\approx 90\%$ of the variance of the data. Fitted to the training dataset,  it involves 80,000 parameters.
    \item {\bf a dense auto-encoder (AE)}: a dense AE operator with
    a 20-dimensional latent representation. The encoder and decoder both involve 3 dense layers with ReLu activations. This AE also amounts to $\approx 90\%$ of the variance of the data. It comprises $\approx$500,000 parameters.
    \item {\bf 2S-CNN model}: a CNN architecture given by (\ref{eq: 2-scale GENN}) 
    with a bilinear parameterization to account for local non-linearities \cite{fablet_bilinear_2018}. It comprises $\approx$55,000 parameters. Importantly, 
    it combines both global and local features through the considered two-scale representation.
\end{itemize}
We let the reader refer to the code provided as supplementary material for more details. 
Given these parameterizations, we investigate different learning strategies and/or optimisation schemes:
\begin{itemize}
    \item {\bf A fixed-step gradient descent}: assuming operator $\Phi$ has been pre-trained, we implement a simple fixed-step gradient descent for variational cost (\ref{eq: var model}). We empirically tune the gradient step and the weighing factors $\lambda_{1,2}$ 
    according to a log-10 scale grid search.  
    We report the best reconstruction performance along the implemented gradient descent pathway. 
    We refer to this optimization scheme as FSGD;
    \item {\bf A fixed-point procedure}: a classic approach to solve for inverse problems using deep learning architectures \cite{chen_learning_2015,liu_image_2018,xie_image_2012} is to train a solver using observation $y$ as inputs. Such approaches may be regarded as the application of a one-step fixed-point solver for criterion (\ref{eq: var model}) \cite{fablet_end--end_2019}.
    The resulting interpolator is  $\Phi(y)$. We refer to this solver as FP(1).
    \item {\bf a learnable gradient-based solver}: the LSTM solver given by (\ref{eq: lstm update}) involves a convolutional LSTM cell whose hidden states involve 5 channels. We refer to this solver as LSTM-S.
\end{itemize}
For benchmarking purposes, we include sparse coding approaches using OMP (Orthogonal Matching Pursuit) and Lasso solutions \cite{mairal_online_2009}. We adapt these schemes for interpolation issues through 
iterative coding-decoding steps, the observed data being kept after each decoding step, except in the final iteration\footnote{These sparse coding schemes are run with MiniBatchDictionaryLearning function in scikitlearn  \cite{pedregosa_scikit-learn_2011}.}. 
Our evaluation relies on three normalized metrics for the test dataset, using $\sigma^2$ 
as normalization factor: the normalized mean square error (NMSE) of the reconstruction over the entire image domain (R-Score), the NMSE of the reconstruction over the unobserved image domain (I-Score) and the NMSE of the projection error $x-\Phi(x)$ for true states $x$ (P-Score).

We report in Tab.\ref{tab:MNIST A} a synthesis of these experiments. We may first notice that the  
learnable solver significantly improves the reconstruction error compared with the direct minimization of (\ref{eq: var model}) given a pretrained model (e.g., 0.54 vs. 1.09 (resp. 0.71 vs. 1.51) for the PCA (resp. AE) version of operator $\Phi$)\footnote{We do not carry a similar experiment with the 2S-CNN architecture as by construction the associated pre-trained version of operator $\Phi$ could lead upon convergence to a meaningless prior $\Phi(x)=x$.}. Despite AE and PCA architectures having very similar representation capabilities for the true states (P-scores of $\approx$ 0.12), the reconstruction performance is much better with a PCA representation, which may relate to the simpler and convex nature of the associated variational cost (\ref{eq: var model}).  
As illustrated in Fig.\ref{fig: descent paths}, the learnable solver identifies a much better descent pathway for the reconstruction error from the differentiation of variational cost (\ref{eq: var model}) than the direct minimization of this cost. 

\begin{figure*}[tb]
    \begin{center}
    \begin{center}
        {\bf Reference}
    \end{center}
    
    \includegraphics[trim={100 50 100 50},clip, width=0.8cm]{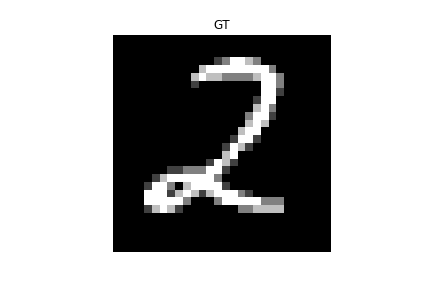}
    \includegraphics[trim={100 50 100 50},clip, width=0.8cm]{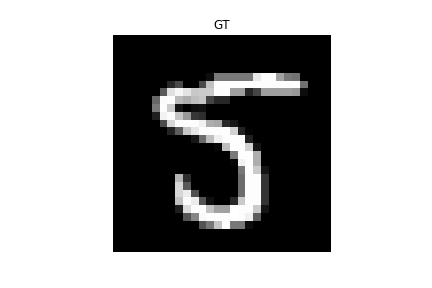}
    \includegraphics[trim={100 50 100 50},clip, width=0.8cm]{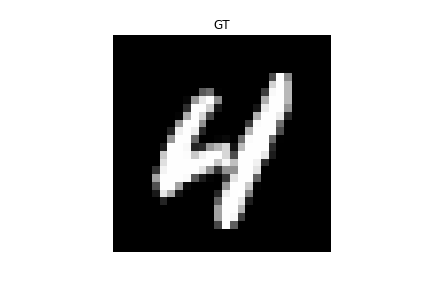}
    \includegraphics[trim={100 50 100 50},clip, width=0.8cm]{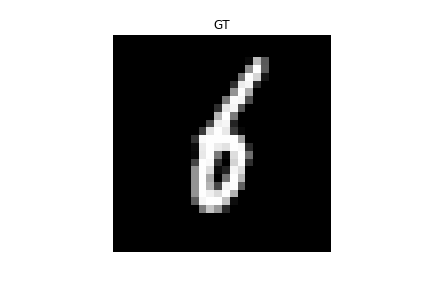}
    \includegraphics[trim={100 50 100 50},clip, width=0.8cm]{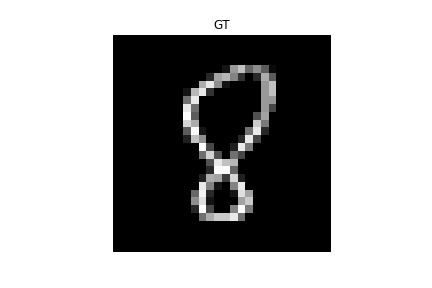}
    \includegraphics[trim={100 50 100 50},clip, width=0.8cm]{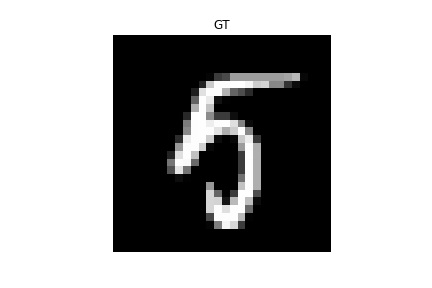}
    \includegraphics[trim={100 50 100 50},clip, width=0.8cm]{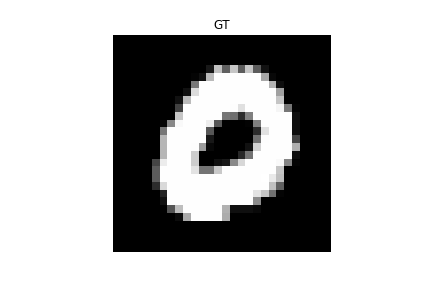}
    \includegraphics[trim={100 50 100 50},clip, width=0.8cm]{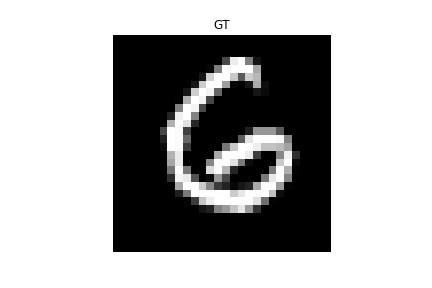}
    \includegraphics[trim={100 50 100 50},clip, width=0.8cm]{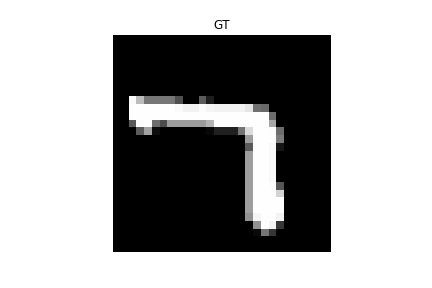}
    \includegraphics[trim={100 50 100 50},clip, width=0.8cm]{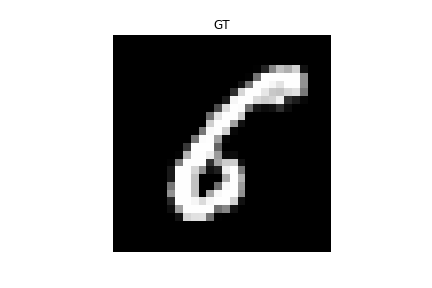}
    \includegraphics[trim={100 50 100 50},clip, width=0.8cm]{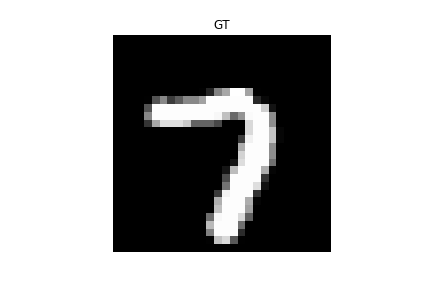}
    \includegraphics[trim={100 50 100 50},clip, width=0.8cm]{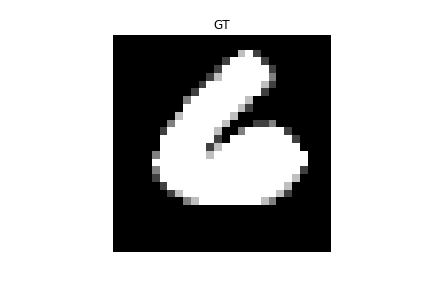}
    \includegraphics[trim={100 50 100 50},clip, width=0.8cm]{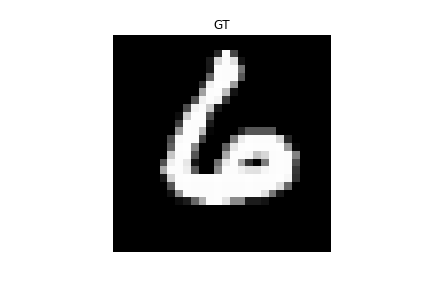}
    \includegraphics[trim={100 50 100 50},clip, width=0.8cm]{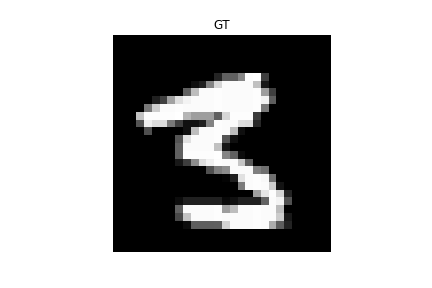}
    \includegraphics[trim={100 50 100 50},clip, width=0.8cm]{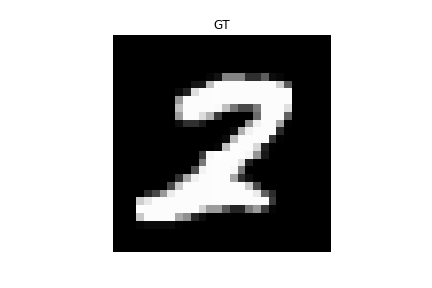}\\
    \vspace*{-0.1cm}
    \begin{center}
        {\bf Observation}
    \end{center}
    \vspace*{-0.15cm}
    
    \includegraphics[trim={100 50 100 50},clip, width=0.8cm]{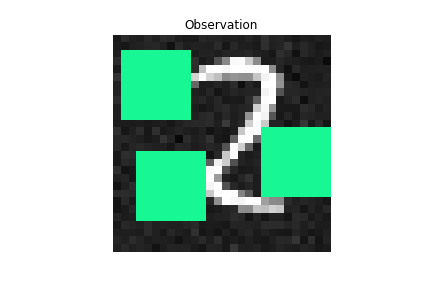}
    \includegraphics[trim={100 50 100 50},clip, width=0.8cm]{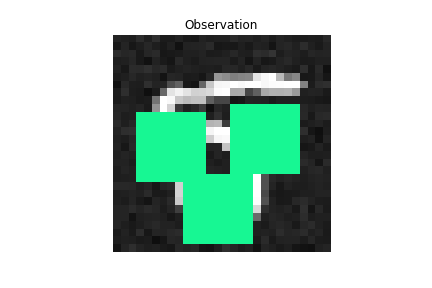}
    \includegraphics[trim={100 50 100 50},clip, width=0.8cm]{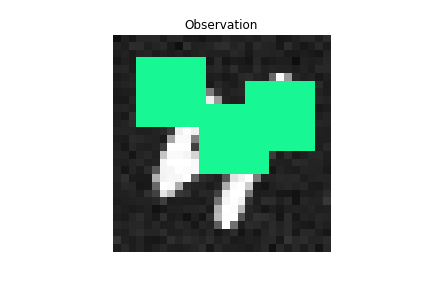}
    \includegraphics[trim={100 50 100 50},clip, width=0.8cm]{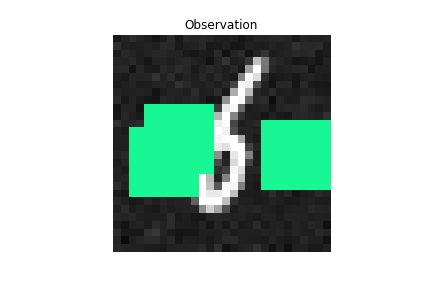}
    \includegraphics[trim={100 50 100 50},clip, width=0.8cm]{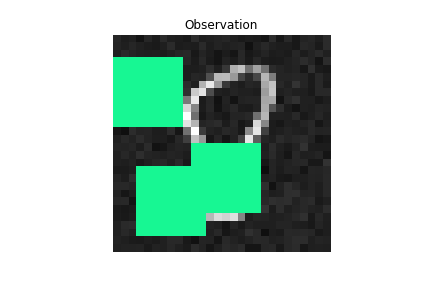}
    \includegraphics[trim={100 50 100 50},clip, width=0.8cm]{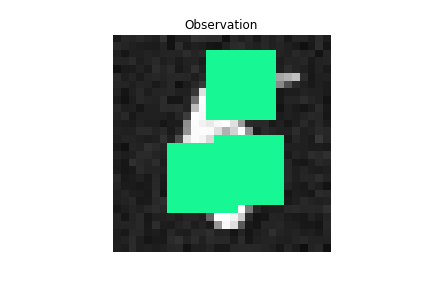}
    \includegraphics[trim={100 50 100 50},clip, width=0.8cm]{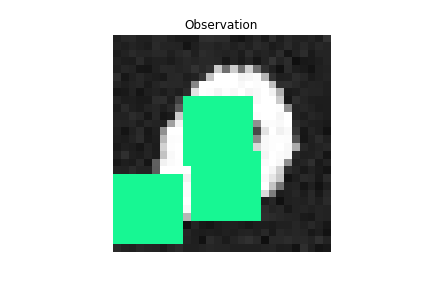}
    \includegraphics[trim={100 50 100 50},clip, width=0.8cm]{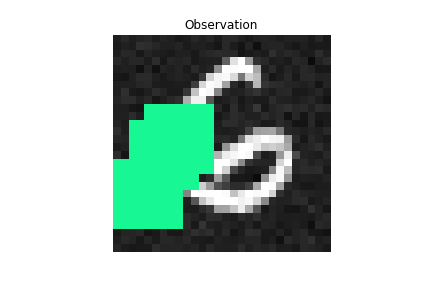}
    \includegraphics[trim={100 50 100 50},clip, width=0.8cm]{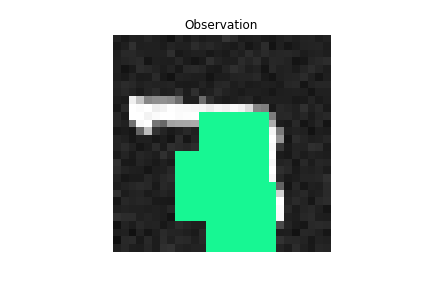}
    \includegraphics[trim={100 50 100 50},clip, width=0.8cm]{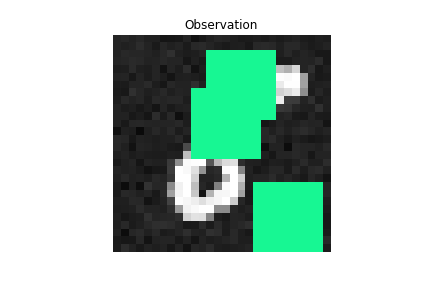}
    \includegraphics[trim={100 50 100 50},clip, width=0.8cm]{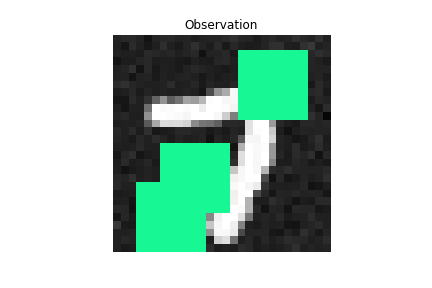}
    \includegraphics[trim={100 50 100 50},clip, width=0.8cm]{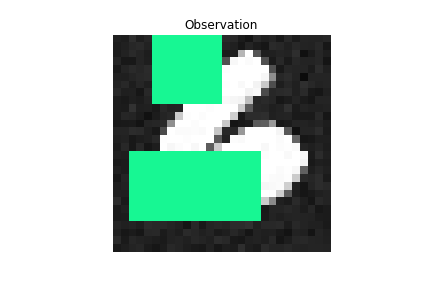}
    \includegraphics[trim={100 50 100 50},clip, width=0.8cm]{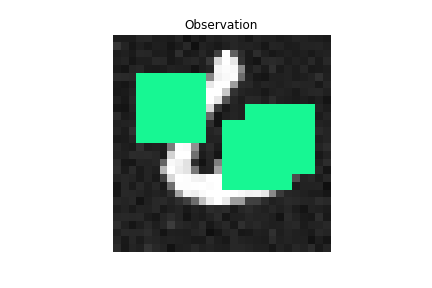}
    \includegraphics[trim={100 50 100 50},clip, width=0.8cm]{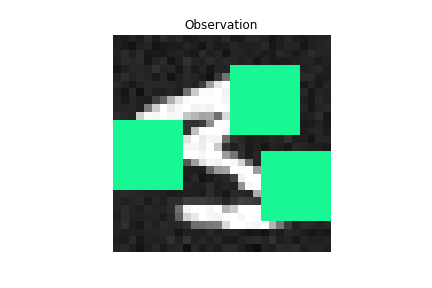}
    \includegraphics[trim={100 50 100 50},clip, width=0.8cm]{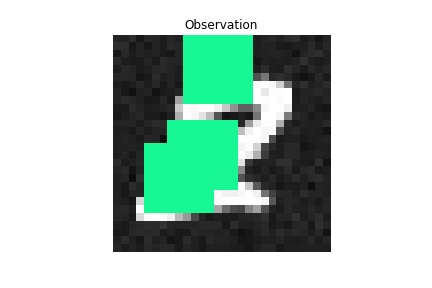}\\

    \vspace*{-0.1cm}
    \begin{center}
        {\bf Sparse coding schemes, adapted from \cite{mairal_online_2009}}
    \end{center}
    \vspace*{-0.15cm}
    \includegraphics[trim={100 50 100 50},clip, width=0.8cm]{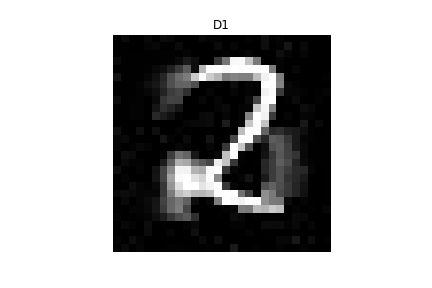}
    \includegraphics[trim={100 50 100 50},clip, width=0.8cm]{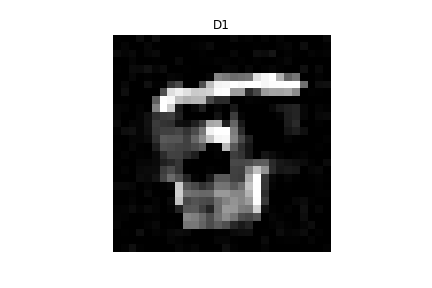}
    \includegraphics[trim={100 50 100 50},clip, width=0.8cm]{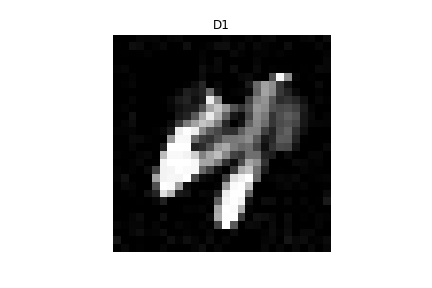}
    \includegraphics[trim={100 50 100 50},clip, width=0.8cm]{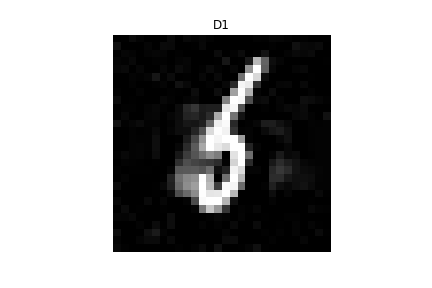}
    \includegraphics[trim={100 50 100 50},clip, width=0.8cm]{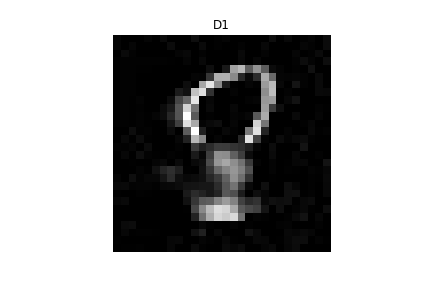}
    \includegraphics[trim={100 50 100 50},clip, width=0.8cm]{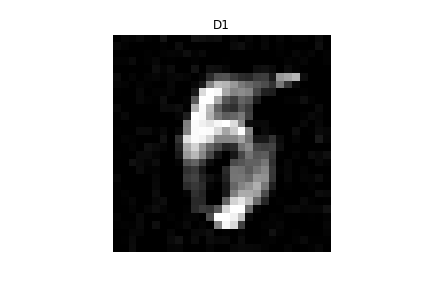}
    \includegraphics[trim={100 50 100 50},clip, width=0.8cm]{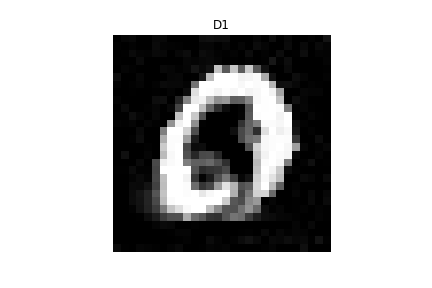}
    \includegraphics[trim={100 50 100 50},clip, width=0.8cm]{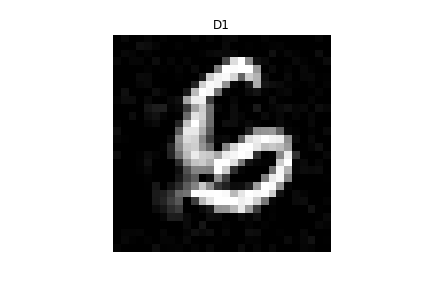}
    \includegraphics[trim={100 50 100 50},clip, width=0.8cm]{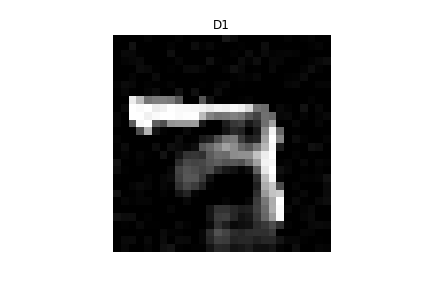}
    \includegraphics[trim={100 50 100 50},clip, width=0.8cm]{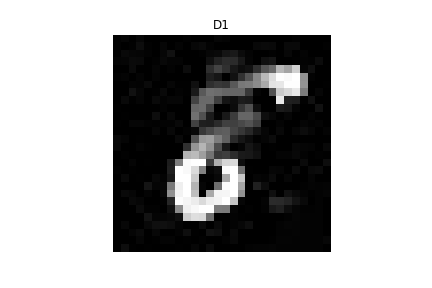}
    \includegraphics[trim={100 50 100 50},clip, width=0.8cm]{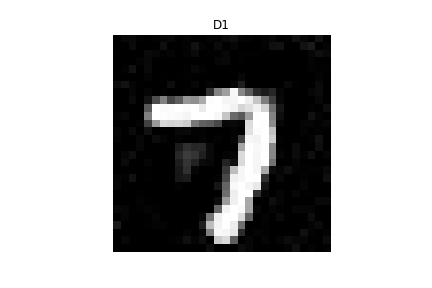}
    \includegraphics[trim={100 50 100 50},clip, width=0.8cm]{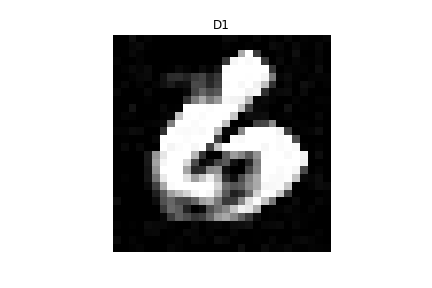}
    \includegraphics[trim={100 50 100 50},clip, width=0.8cm]{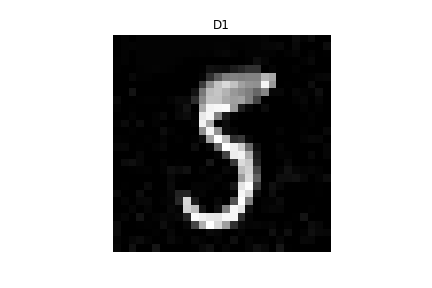}
    \includegraphics[trim={100 50 100 50},clip, width=0.8cm]{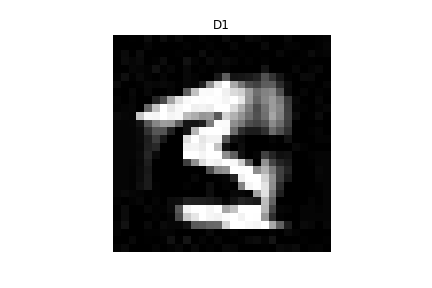}
    \includegraphics[trim={100 50 100 50},clip, width=0.8cm]{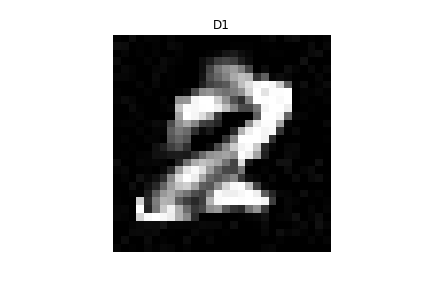}\\

    \includegraphics[trim={100 50 100 50},clip, width=0.8cm]{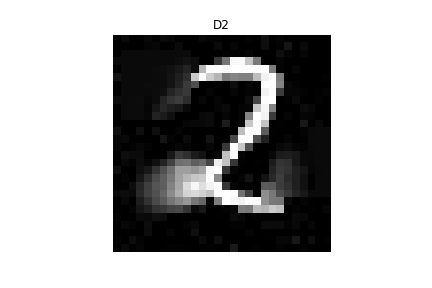}
    \includegraphics[trim={100 50 100 50},clip, width=0.8cm]{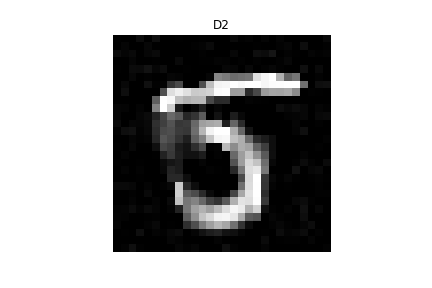}
    \includegraphics[trim={100 50 100 50},clip, width=0.8cm]{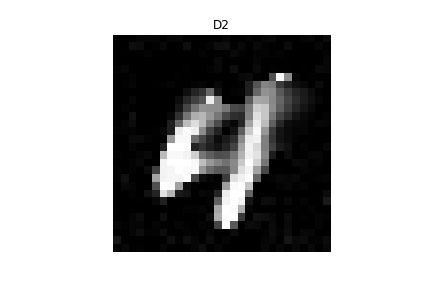}
    \includegraphics[trim={100 50 100 50},clip, width=0.8cm]{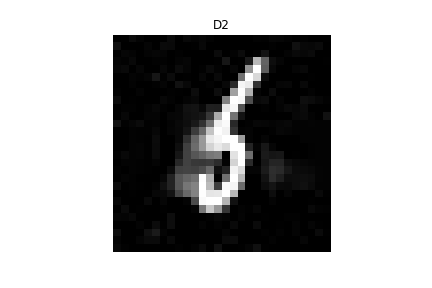}
    \includegraphics[trim={100 50 100 50},clip, width=0.8cm]{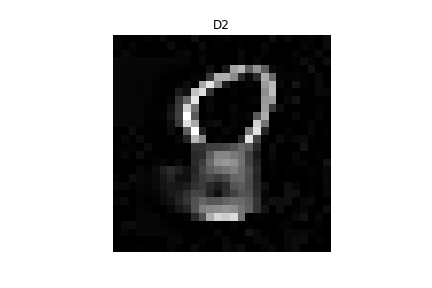}
    \includegraphics[trim={100 50 100 50},clip, width=0.8cm]{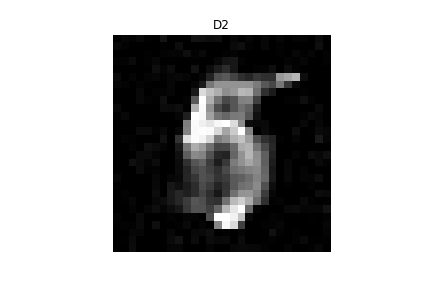}
    \includegraphics[trim={100 50 100 50},clip, width=0.8cm]{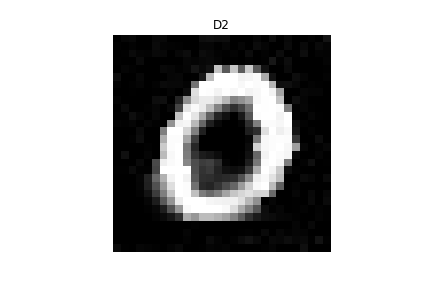}
    \includegraphics[trim={100 50 100 50},clip, width=0.8cm]{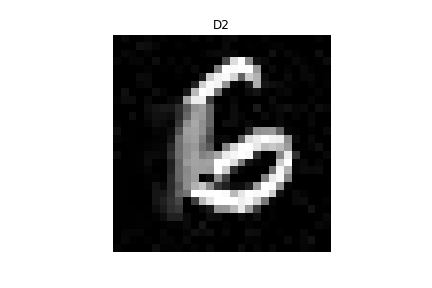}
    \includegraphics[trim={100 50 100 50},clip, width=0.8cm]{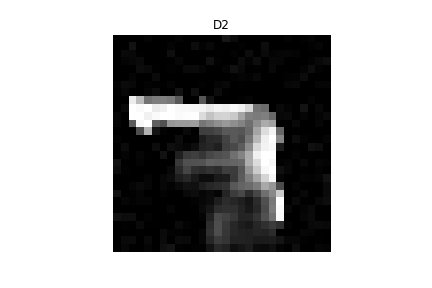}
    \includegraphics[trim={100 50 100 50},clip, width=0.8cm]{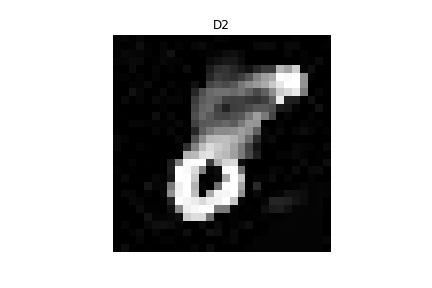}
    \includegraphics[trim={100 50 100 50},clip, width=0.8cm]{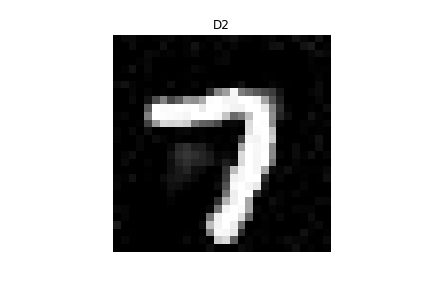}
    \includegraphics[trim={100 50 100 50},clip, width=0.8cm]{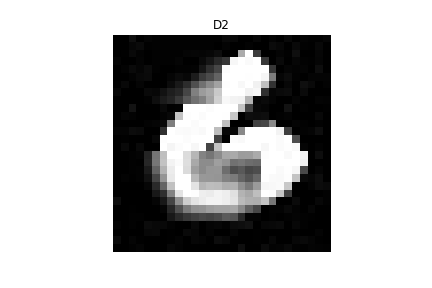}
    \includegraphics[trim={100 50 100 50},clip, width=0.8cm]{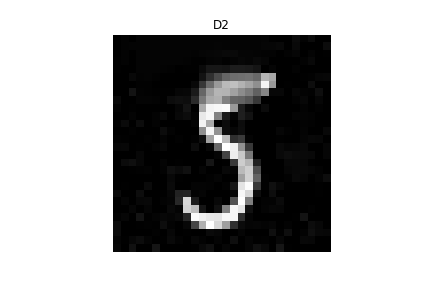}
    \includegraphics[trim={100 50 100 50},clip, width=0.8cm]{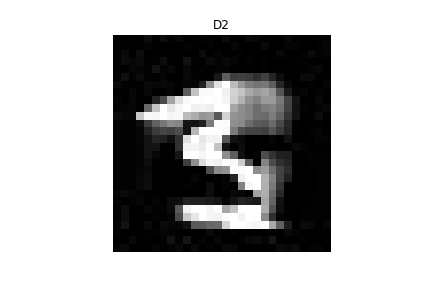}
    \includegraphics[trim={100 50 100 50},clip, width=0.8cm]{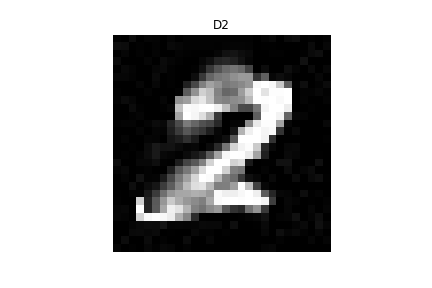}\\

    \vspace*{-0.1cm}
    \begin{center}
        {\bf PCA-based Operator $\Phi$}
    \end{center}
    \vspace*{-0.15cm}
   \includegraphics[trim={100 50 100 50},clip, width=0.8cm]{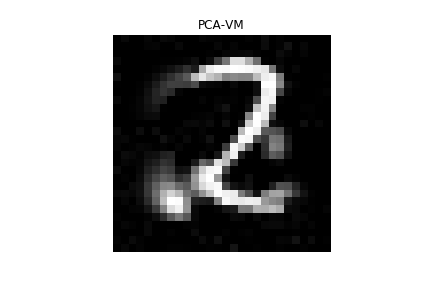}
    \includegraphics[trim={100 50 100 50},clip, width=0.8cm]{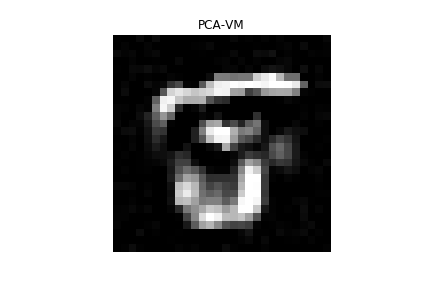}
    \includegraphics[trim={100 50 100 50},clip, width=0.8cm]{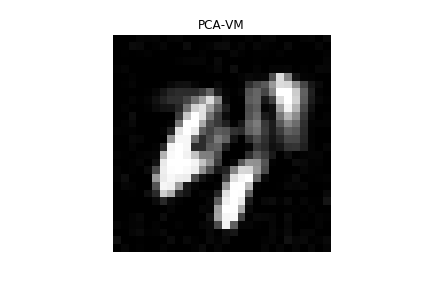}
    \includegraphics[trim={100 50 100 50},clip, width=0.8cm]{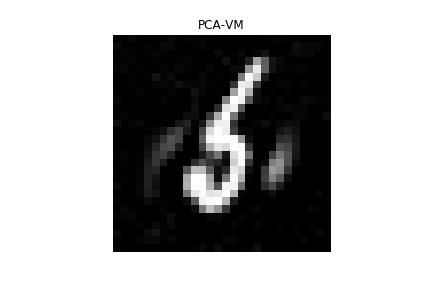}
    \includegraphics[trim={100 50 100 50},clip, width=0.8cm]{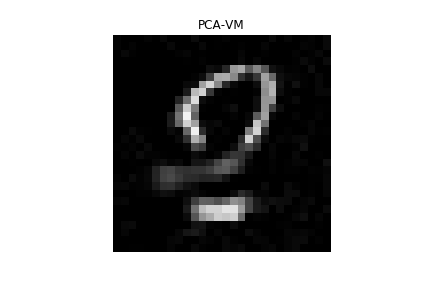}
    \includegraphics[trim={100 50 100 50},clip, width=0.8cm]{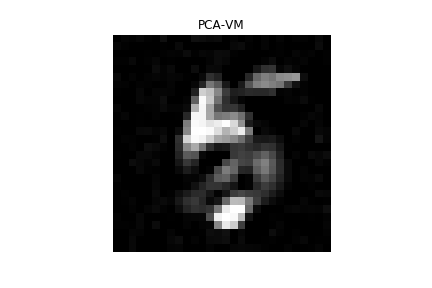}
    \includegraphics[trim={100 50 100 50},clip, width=0.8cm]{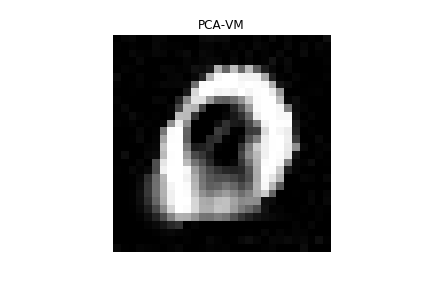}
    \includegraphics[trim={100 50 100 50},clip, width=0.8cm]{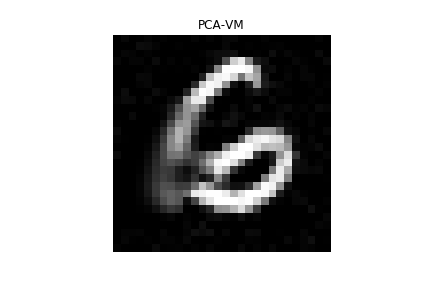}
    \includegraphics[trim={100 50 100 50},clip, width=0.8cm]{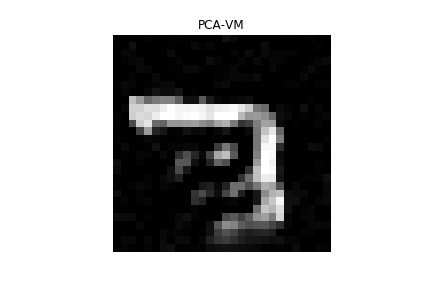}
    \includegraphics[trim={100 50 100 50},clip, width=0.8cm]{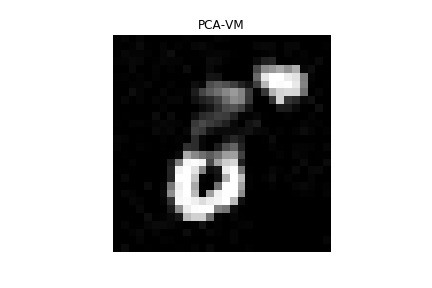}
    \includegraphics[trim={100 50 100 50},clip, width=0.8cm]{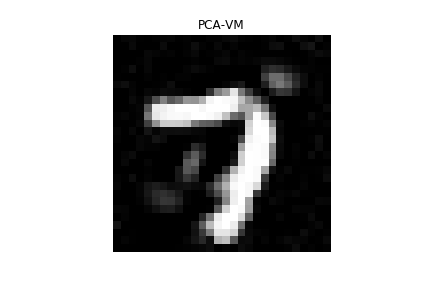}
    \includegraphics[trim={100 50 100 50},clip, width=0.8cm]{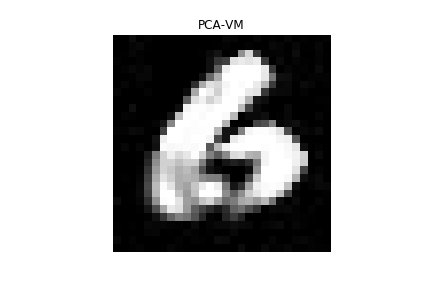}
    \includegraphics[trim={100 50 100 50},clip, width=0.8cm]{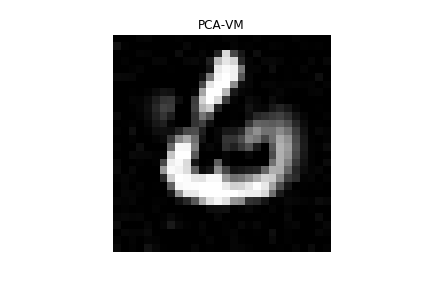}
    \includegraphics[trim={100 50 100 50},clip, width=0.8cm]{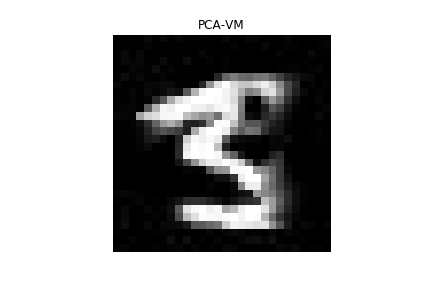}
    \includegraphics[trim={100 50 100 50},clip, width=0.8cm]{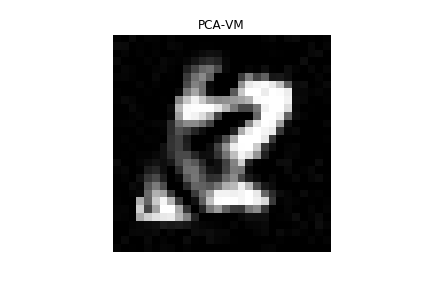}\\

    \includegraphics[trim={100 50 100 50},clip, width=0.8cm]{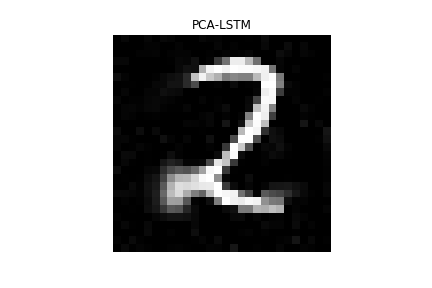}
    \includegraphics[trim={100 50 100 50},clip, width=0.8cm]{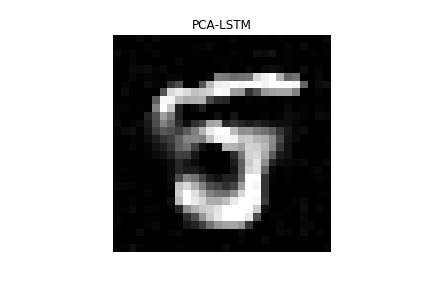}
    \includegraphics[trim={100 50 100 50},clip, width=0.8cm]{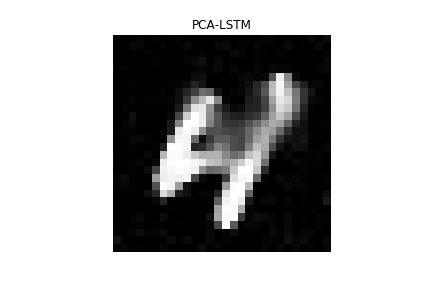}
    \includegraphics[trim={100 50 100 50},clip, width=0.8cm]{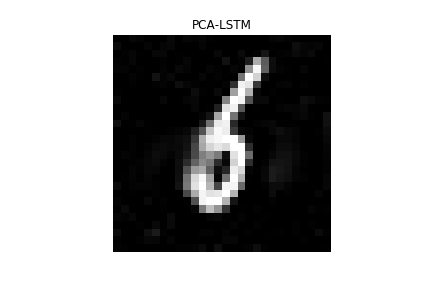}
    \includegraphics[trim={100 50 100 50},clip, width=0.8cm]{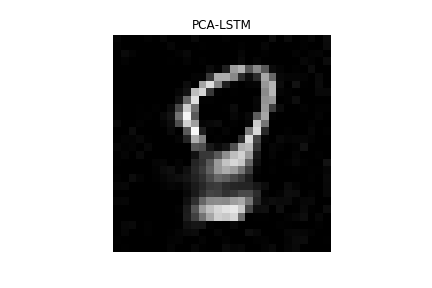}
    \includegraphics[trim={100 50 100 50},clip, width=0.8cm]{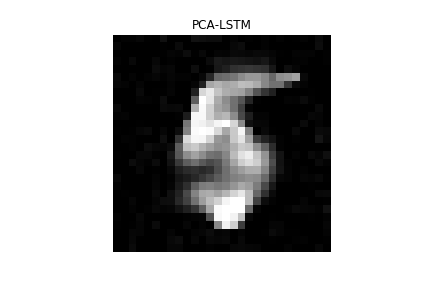}
    \includegraphics[trim={100 50 100 50},clip, width=0.8cm]{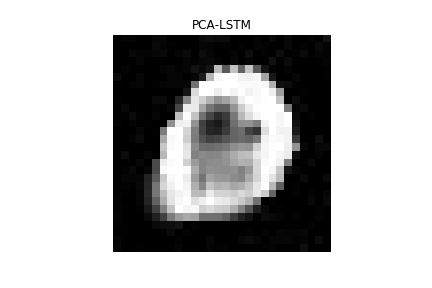}
    \includegraphics[trim={100 50 100 50},clip, width=0.8cm]{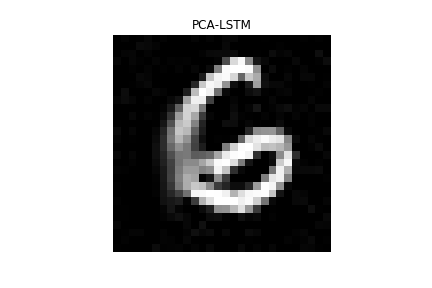}
    \includegraphics[trim={100 50 100 50},clip, width=0.8cm]{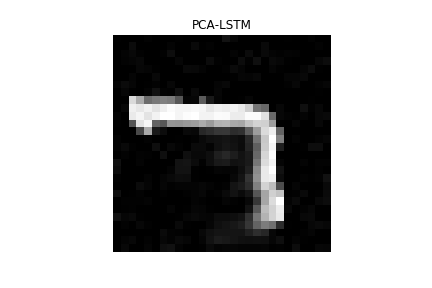}
    \includegraphics[trim={100 50 100 50},clip, width=0.8cm]{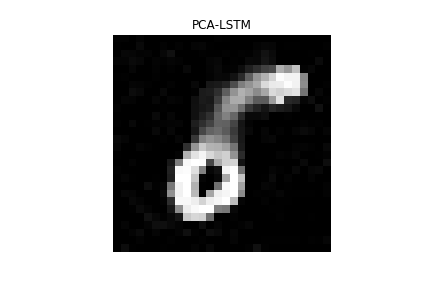}
    \includegraphics[trim={100 50 100 50},clip, width=0.8cm]{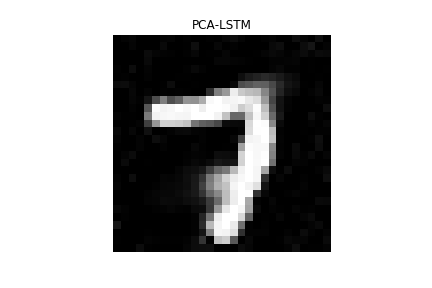}
    \includegraphics[trim={100 50 100 50},clip, width=0.8cm]{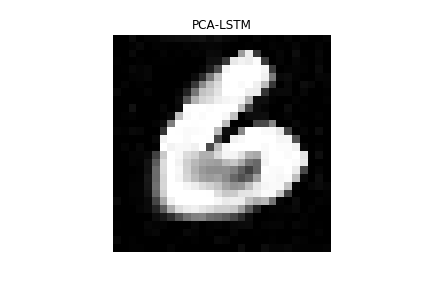}
    \includegraphics[trim={100 50 100 50},clip, width=0.8cm]{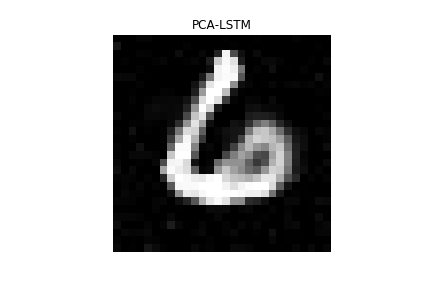}
    \includegraphics[trim={100 50 100 50},clip, width=0.8cm]{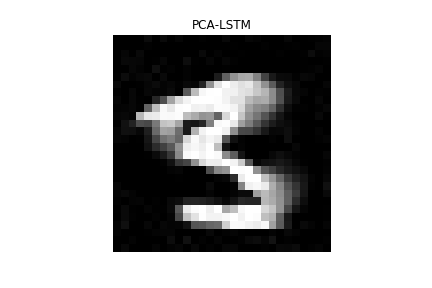}
    \includegraphics[trim={100 50 100 50},clip, width=0.8cm]{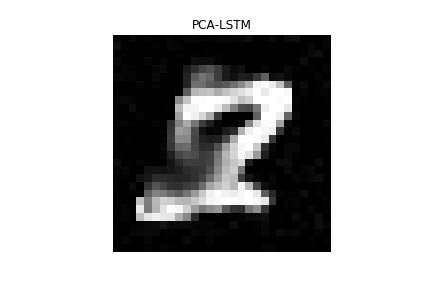}\\

    \vspace*{-0.1cm}
    \begin{center}
        {\bf AE-based Operator $\Phi$}
    \end{center}
    \vspace*{-0.15cm}
    \includegraphics[trim={100 50 100 50},clip, width=0.8cm]{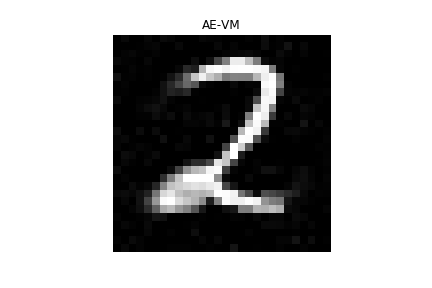}
    \includegraphics[trim={100 50 100 50},clip, width=0.8cm]{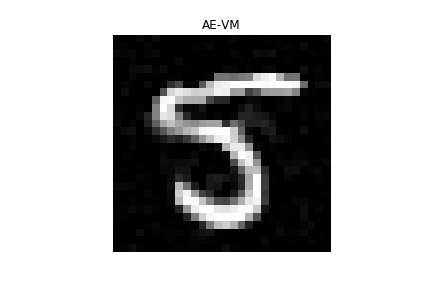}
    \includegraphics[trim={100 50 100 50},clip, width=0.8cm]{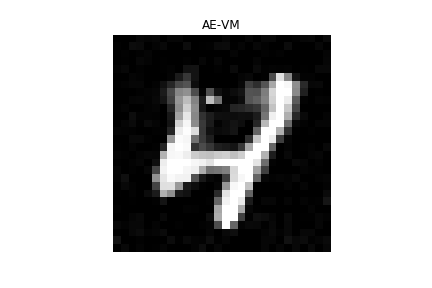}
    \includegraphics[trim={100 50 100 50},clip, width=0.8cm]{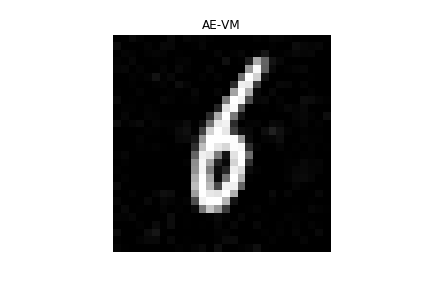}
    \includegraphics[trim={100 50 100 50},clip, width=0.8cm]{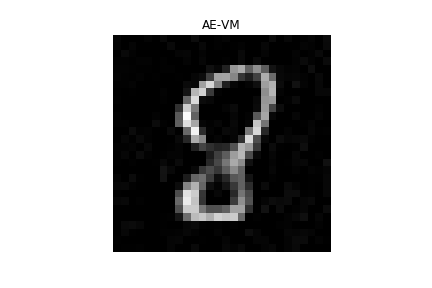}
    \includegraphics[trim={100 50 100 50},clip, width=0.8cm]{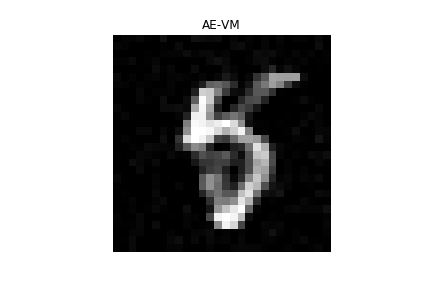}
    \includegraphics[trim={100 50 100 50},clip, width=0.8cm]{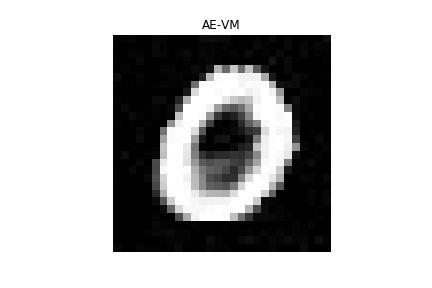}
    \includegraphics[trim={100 50 100 50},clip, width=0.8cm]{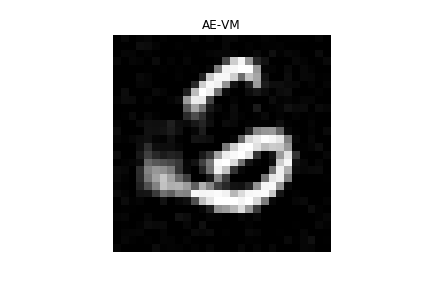}
    \includegraphics[trim={100 50 100 50},clip, width=0.8cm]{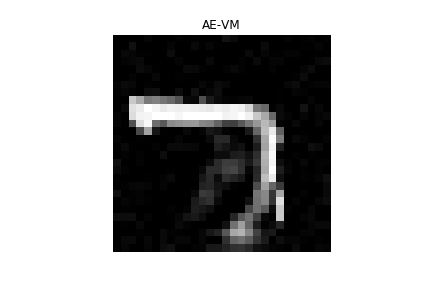}
    \includegraphics[trim={100 50 100 50},clip, width=0.8cm]{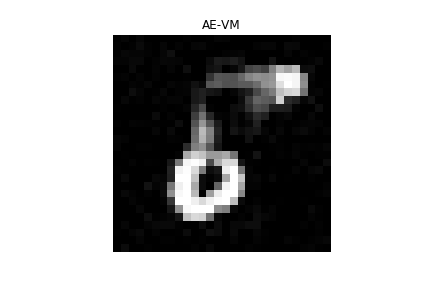}
    \includegraphics[trim={100 50 100 50},clip, width=0.8cm]{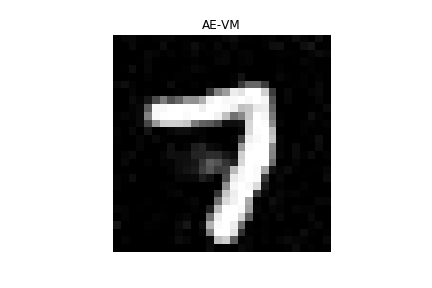}
    \includegraphics[trim={100 50 100 50},clip, width=0.8cm]{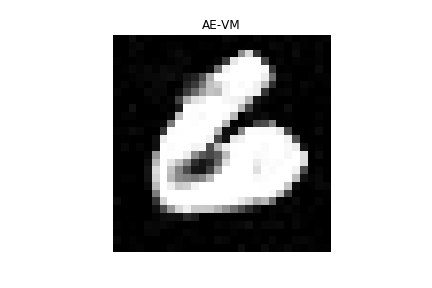}
    \includegraphics[trim={100 50 100 50},clip, width=0.8cm]{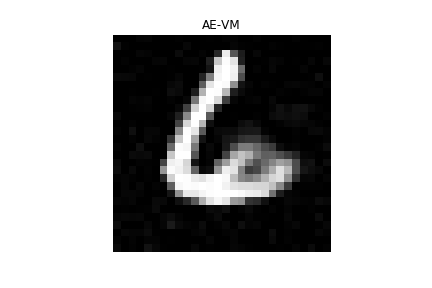}
    \includegraphics[trim={100 50 100 50},clip, width=0.8cm]{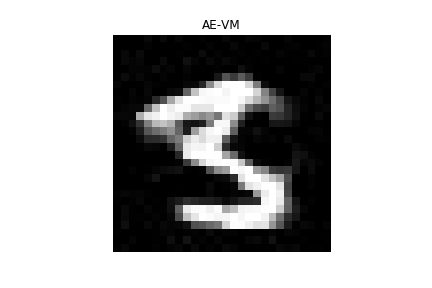}
    \includegraphics[trim={100 50 100 50},clip, width=0.8cm]{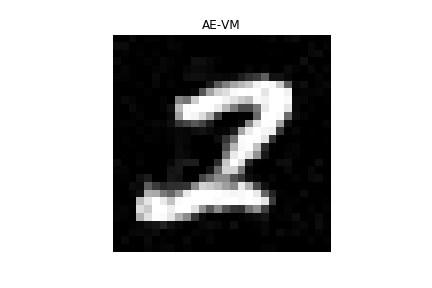}\\

    \includegraphics[trim={100 50 100 50},clip, width=0.8cm]{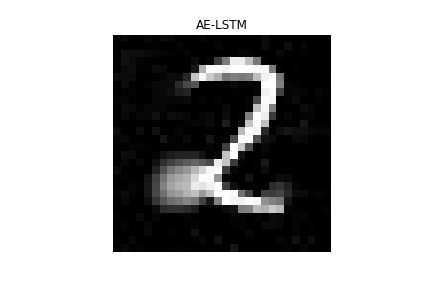}
    \includegraphics[trim={100 50 100 50},clip, width=0.8cm]{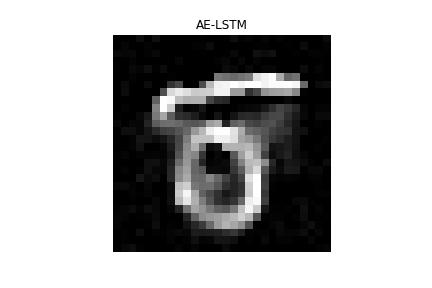}
    \includegraphics[trim={100 50 100 50},clip, width=0.8cm]{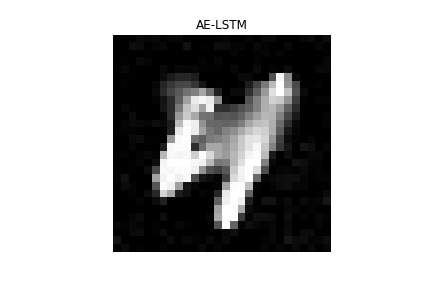}
    \includegraphics[trim={100 50 100 50},clip, width=0.8cm]{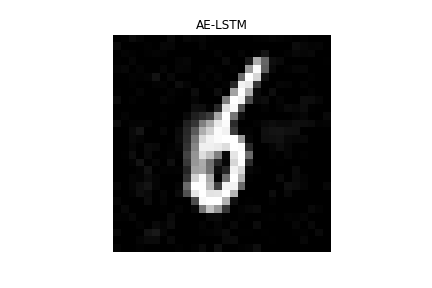}
    \includegraphics[trim={100 50 100 50},clip, width=0.8cm]{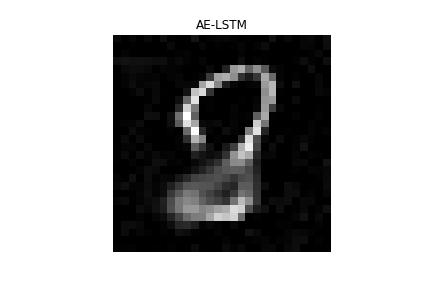}
    \includegraphics[trim={100 50 100 50},clip, width=0.8cm]{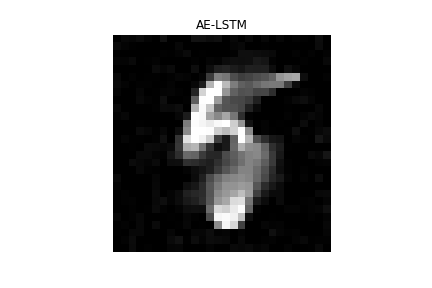}
    \includegraphics[trim={100 50 100 50},clip, width=0.8cm]{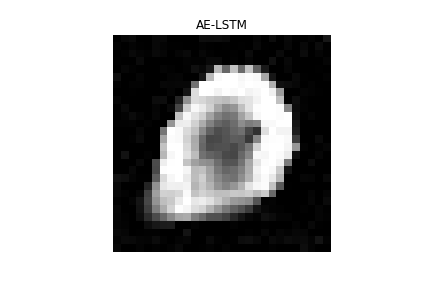}
    \includegraphics[trim={100 50 100 50},clip, width=0.8cm]{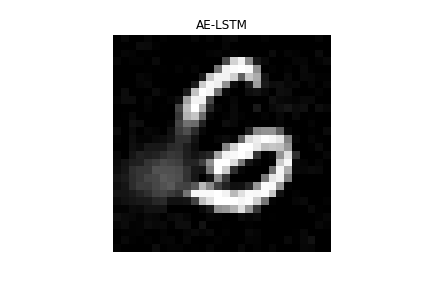}
    \includegraphics[trim={100 50 100 50},clip, width=0.8cm]{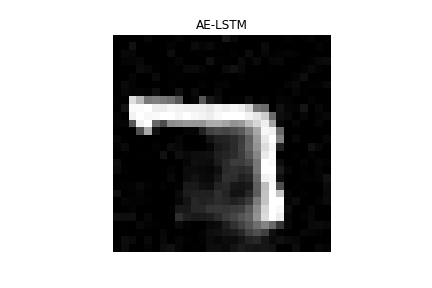}
    \includegraphics[trim={100 50 100 50},clip, width=0.8cm]{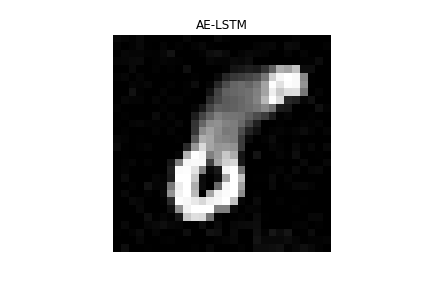}
    \includegraphics[trim={100 50 100 50},clip, width=0.8cm]{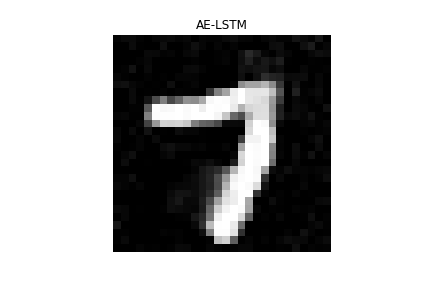}
    \includegraphics[trim={100 50 100 50},clip, width=0.8cm]{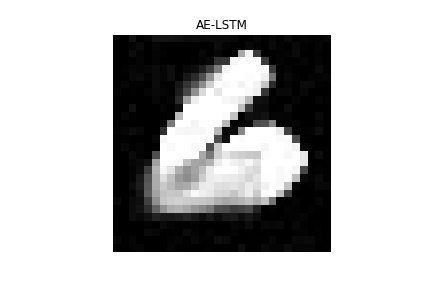}
    \includegraphics[trim={100 50 100 50},clip, width=0.8cm]{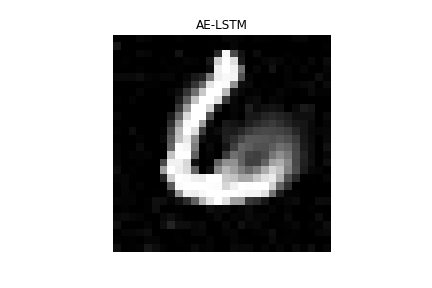}
    \includegraphics[trim={100 50 100 50},clip, width=0.8cm]{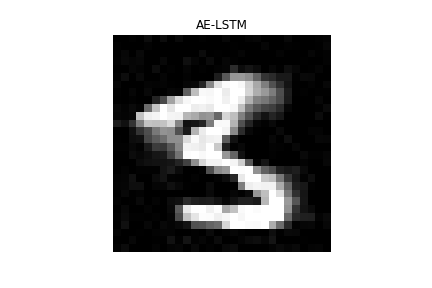}
    \includegraphics[trim={100 50 100 50},clip, width=0.8cm]{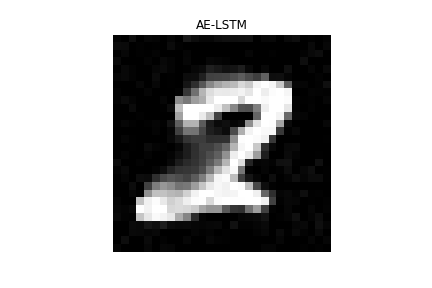}\\

    \vspace*{-0.10cm}
    \begin{center}
        {\bf 2S-CNN-based Operator $\Phi$}
    \end{center}
    \vspace*{-0.15cm}
    \includegraphics[trim={100 50 100 50},clip, width=0.8cm]{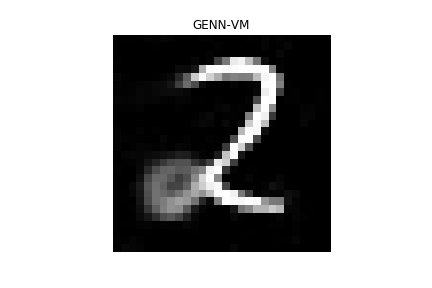}
    \includegraphics[trim={100 50 100 50},clip, width=0.8cm]{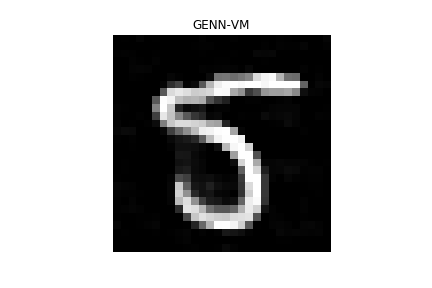}
    \includegraphics[trim={100 50 100 50},clip, width=0.8cm]{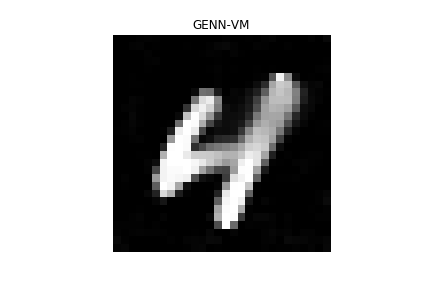}
    \includegraphics[trim={100 50 100 50},clip, width=0.8cm]{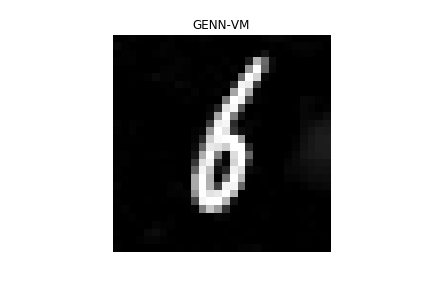}
    \includegraphics[trim={100 50 100 50},clip, width=0.8cm]{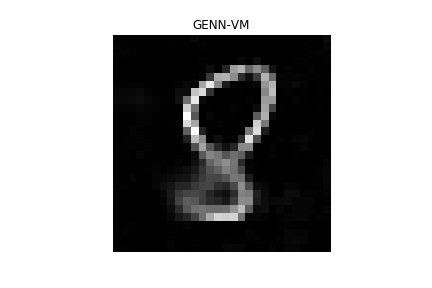}
    \includegraphics[trim={100 50 100 50},clip, width=0.8cm]{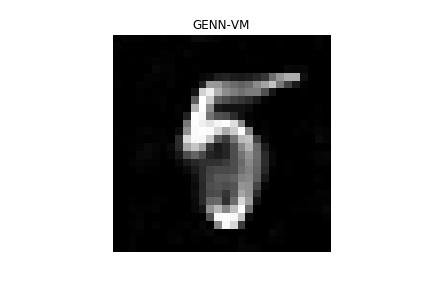}
    \includegraphics[trim={100 50 100 50},clip, width=0.8cm]{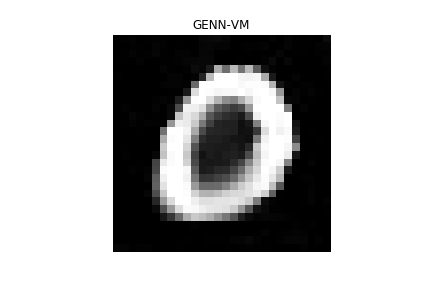}
    \includegraphics[trim={100 50 100 50},clip, width=0.8cm]{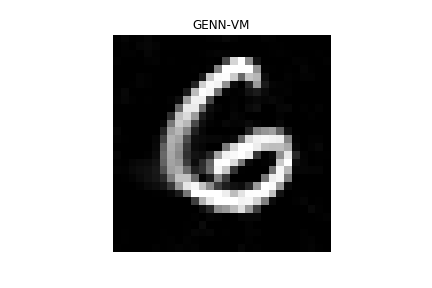}
    \includegraphics[trim={100 50 100 50},clip, width=0.8cm]{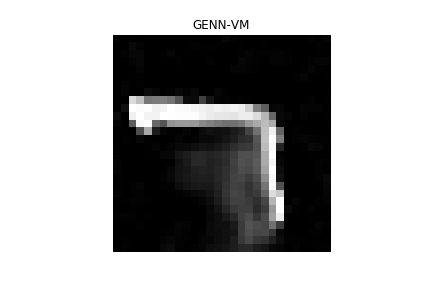}
    \includegraphics[trim={100 50 100 50},clip, width=0.8cm]{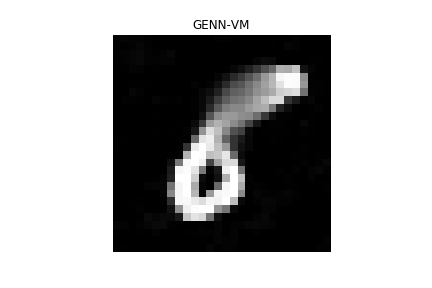}
    \includegraphics[trim={100 50 100 50},clip, width=0.8cm]{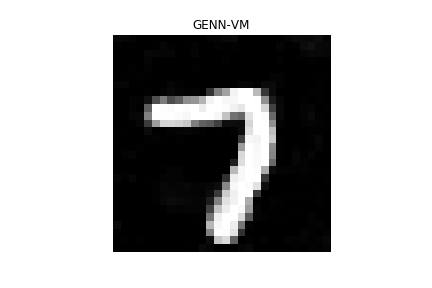}
    \includegraphics[trim={100 50 100 50},clip, width=0.8cm]{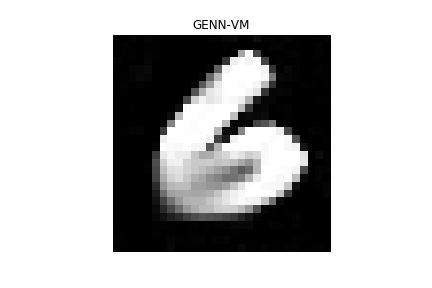}
    \includegraphics[trim={100 50 100 50},clip, width=0.8cm]{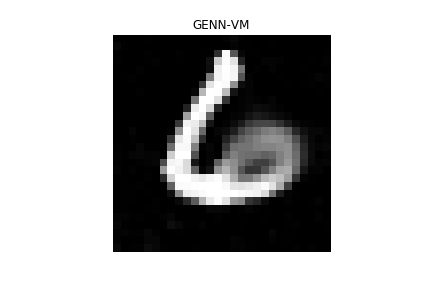}
    \includegraphics[trim={100 50 100 50},clip, width=0.8cm]{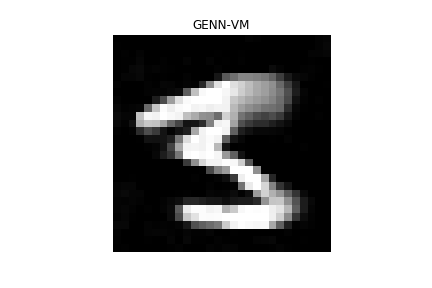}
    \includegraphics[trim={100 50 100 50},clip, width=0.8cm]{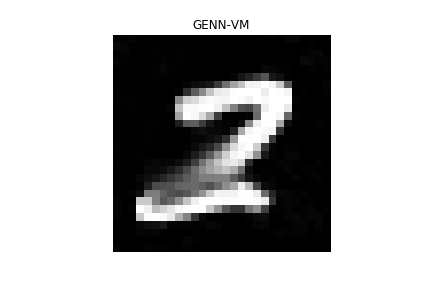}\\

    \includegraphics[trim={100 50 100 50},clip, width=0.8cm]{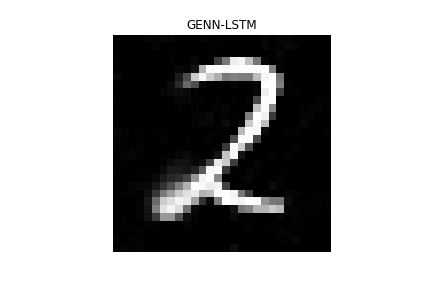}
    \includegraphics[trim={100 50 100 50},clip, width=0.8cm]{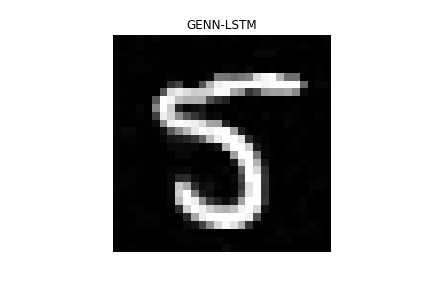}
    \includegraphics[trim={100 50 100 50},clip, width=0.8cm]{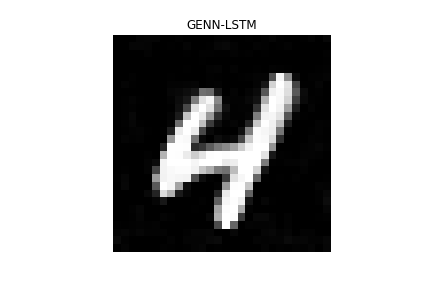}
    \includegraphics[trim={100 50 100 50},clip, width=0.8cm]{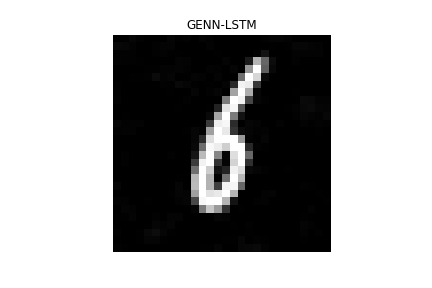}
    \includegraphics[trim={100 50 100 50},clip, width=0.8cm]{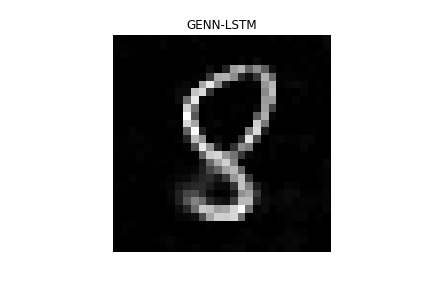}
    \includegraphics[trim={100 50 100 50},clip, width=0.8cm]{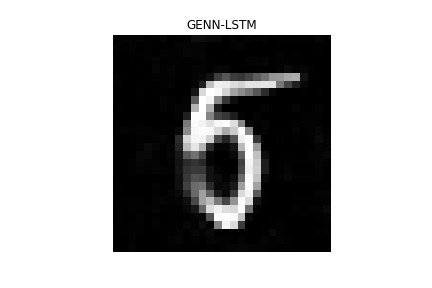}
    \includegraphics[trim={100 50 100 50},clip, width=0.8cm]{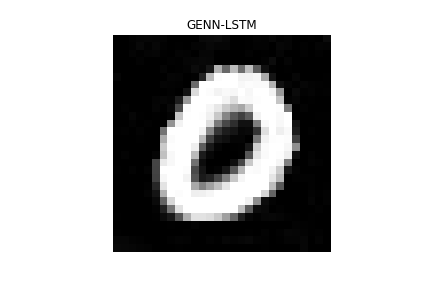}
    \includegraphics[trim={100 50 100 50},clip, width=0.8cm]{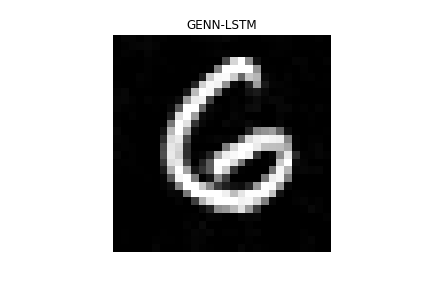}
    \includegraphics[trim={100 50 100 50},clip, width=0.8cm]{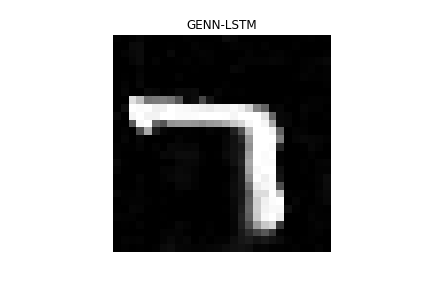}
    \includegraphics[trim={100 50 100 50},clip, width=0.8cm]{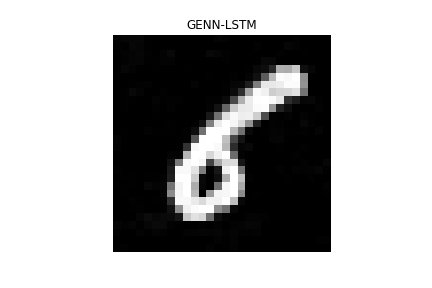}
    \includegraphics[trim={100 50 100 50},clip, width=0.8cm]{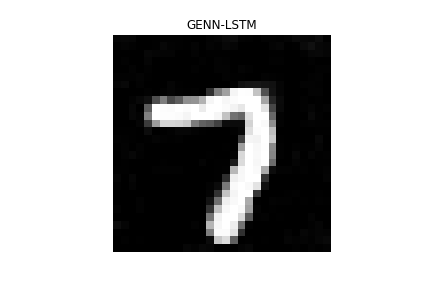}
    \includegraphics[trim={100 50 100 50},clip, width=0.8cm]{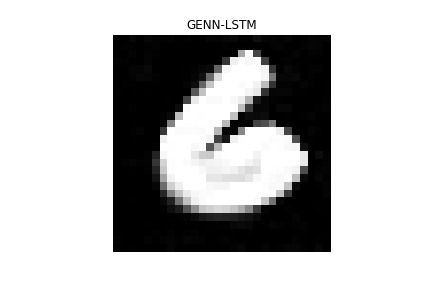}
    \includegraphics[trim={100 50 100 50},clip, width=0.8cm]{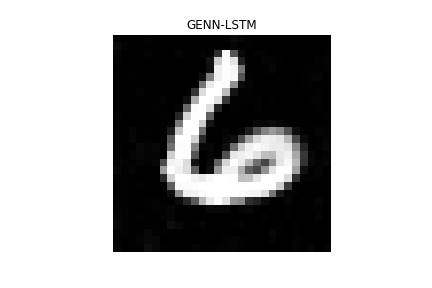}
    \includegraphics[trim={100 50 100 50},clip, width=0.8cm]{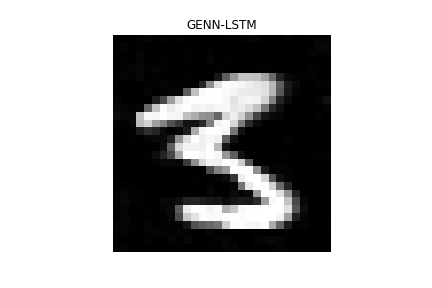}
    \includegraphics[trim={100 50 100 50},clip, width=0.8cm]{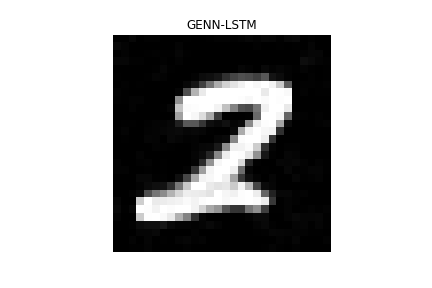}\\
    \end{center}

    \caption{{\bf Reconstruction examples for MNIST experiment}. We randomly sample 15 examples for MNIST test dataset: first row, reference images; second row, observed images; third and fourth rows, sparse decomposition methods \cite{mairal_online_2009} using respectively OMP and Lasso solutions.
    From the fifth row to the tenth row, we report results using the proposed framework using  PCA-based, AE-based and 2S-CNN-based parameterizations for operator $\Phi$. For each example, we first report the output of a gradient descent of the considered variational cost (e.g., the fifth row  for the PCA-based setting) and of the learnt LSTM solver (e.g., the sixth row for the PCA-based setting). We let the reader refer to the main text for the details on the considered experimental setting and parameterizations.}
    \label{fig: mnist exples}
\end{figure*}

The second key result is the significant gain issued from the joint learning of operator $\Phi$ and of the solver (0.33 vs. 0.54 for the best I-scores of 2S-CNN and PCA representations). Note that the 2S-CNN-based operator $\Phi$ used as a direct solution for the inversion (FP(1) solver in Tab.\ref{tab:MNIST A}) is also significantly outperformed. Intriguingly, the better reconstruction performance issued from the 2S-CNN representation involves a worse P-score. It suggests that a coarser approximation of true states by operator $\Phi$ may be better suited to reconstruction issues than the one best describing the true states. The visual inspection of examples in Fig.\ref{fig: mnist exples} is in line with these results. The 2S-CNN parameterization in the joint learning setting clearly outperforms the other approaches. The FSGD minimization of the learned 2S-CNN-based variational cost leads to visually relevant reconstructions,
though the contrast of interpolated areas is often too low.
This, with the consistent descent pathways of the FSGD and learnable solvers for the 2S-CNN case in
Fig.\ref{fig: descent paths}, supports the consistency of the learned variational representation. Fig.\ref{fig: descent paths} also stresses the fast convergence of the learnable solvers with 10 to 20 iterations, compared with hundreds or thousands for the FSGD scheme.

\subsection{Multivariate signals governed by ODEs}

The second case-study addresses multivariate signals
governed by ODEs so that we are provided with a theoretical ground-truth for the true representation of the data: namely,  Lorenz-63 and Lorenz-96 dynamics, which are widely used for benchmarking experiments in geoscience and data assimilation. Both systems involve bilinear ODEs which lead to chaotic patterns under the considered parametrizations. We detail our experimental setting, which involves noisy and undersampled observations, in the Supplementary Material.

We consider two architectures for operator $\Phi$. A Neural ODE representation is first fully-informed by the true ODE using a bilinear ResNet architecture\footnote{It relies on a 4$^{th}$-order Runge-Kutta integration which provides a good approximation of the integration scheme considered for data generation.} \cite{fablet_bilinear_2018}, referred to as L63-RK4 and L96-RK4. We also investigate CNN-based representations (\ref{eq: 2-scale GENN}), where operators $\Phi_{1,2}$ involve bilinear convolutional blocks with 30 channels. Referred to as L63-2S-CNN and L96-2S-CNN, they involve respectively 15,000 and 52,000 parameters.

We consider the same optimization schemes as for MNIST case-study. Regarding evaluation metrics, we focus on the R-score and the P-score\footnote{Given the noise level and regular sampling, R-score and I-score provide relatively similar information.}. Tab.\ref{tab: L63/L96} further supports the relevance of the proposed framework. Similarly to MNIST case-study, the learnable LSTM solver clearly outperforms the FSGD minimization.  The joint learning of the variational representation and of the solver further improves the reconstruction performance. Besides, we also retrieve that the best reconstruction schemes involve a significantly worse representation performance for the true states.

\begin{table*}[tb]
    \footnotesize
    \centering
    \begin{tabular}{|C{2.5cm}||C{1.5cm}|C{2cm}|C{1.75cm}|C{1.75cm}|}
    \toprule
    \toprule
     \bf Model $\Phi$ &\bf Joint learning &\bf Solver&\bf R-Score&\bf P-score\\
    \toprule
    \toprule
      {\bf L63-RK4}& No& FSGD & 2.71e-3&{\bf 1.41e-7}\\    
     & No & LSTM-S & 1.52e-3 &{\bf  1.41e-7}\\
     {\bf L63-2S-CNN} & Yes & LSTM-S & \bf 1.00e-3 & 2.78e-4 \\
    \toprule
    \toprule
      {\bf L96-RK4}& No& FSGD & 5.65e-2&{\bf  6.70e-7}\\    
     & No & LSTM-S & 4.46e-2&{\bf 6.70e-7}\\
     {\bf L96-2S-CNN} & Yes & LSTM-S & \bf 1.94e-2& 3.70e-3  \\
    \bottomrule
    \bottomrule
    \end{tabular}
    \caption{{\bf Experiment with multivariate signals governed by ODEs:} we report for Lorenz-63 and Lorenz-96 dynamics  (\ref{eq:lorenz-63}-\ref{eq:lorenz-96}) reconstruction experiments from noisy and under-sampled observations. We let the reader refer to the main text for the details on the experimental setting and the parameterizations considered for operator $\Phi$ in (\ref{eq: var model}). The evaluation procedure is similar to Tab.\ref{tab:MNIST A}.    }
    \label{tab: L63/L96}
\end{table*}

\begin{figure*}
\begin{center}
 \includegraphics[height=4.25cm]{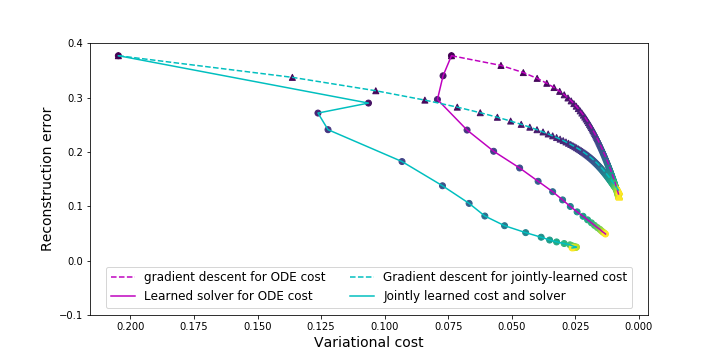}
\includegraphics[height=4.25cm]{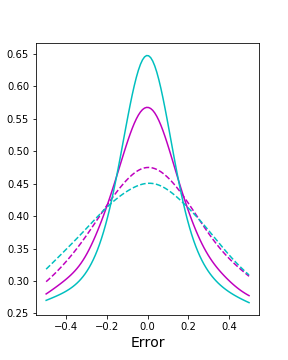} \\  
\end{center}
\begin{tabular}{c||ccc}
    {\footnotesize {\bf True and observed states}} &
    \multicolumn{3}{c}{{\footnotesize {\bf Reconstruction examples and associated error maps}}} \\
    \includegraphics[trim={55 110 150 110},clip, width=3cm]{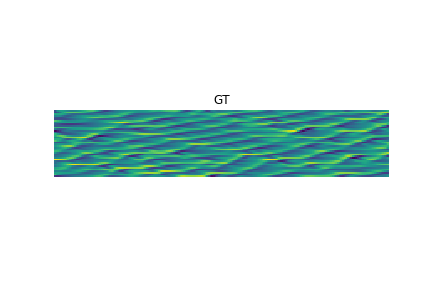}&
    \includegraphics[trim={55 110 150 110},clip, width=3cm]{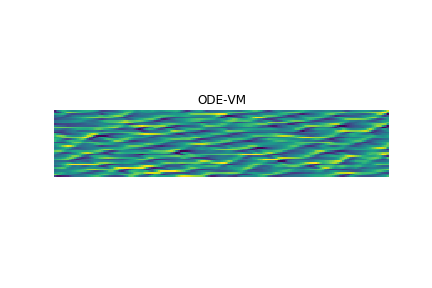}&
    \includegraphics[trim={55 110 150 110},clip, width=3cm]{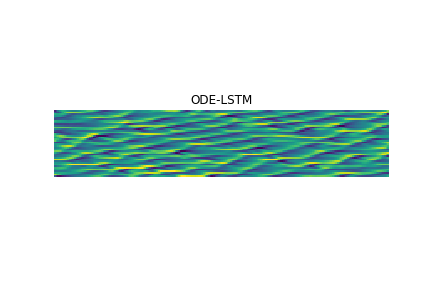}&
    \includegraphics[trim={55 110 150 110},clip, width=3cm]{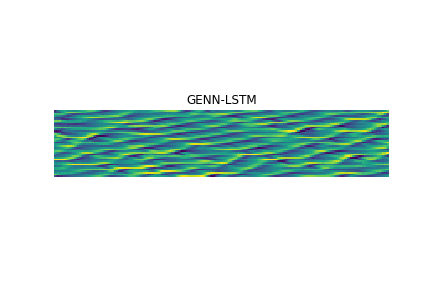}\\
    
    \includegraphics[trim={55 110 150 110},clip, width=3cm]{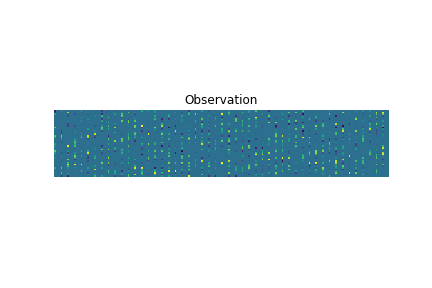}&
    \includegraphics[trim={55 110 150 110},clip, width=3cm]{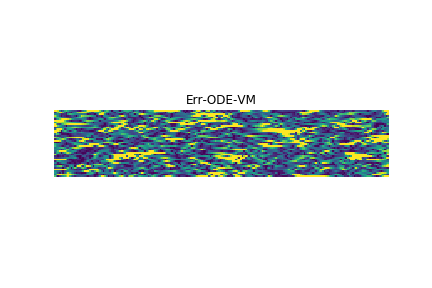}&
    \includegraphics[trim={55 110 150 110},clip, width=3cm]{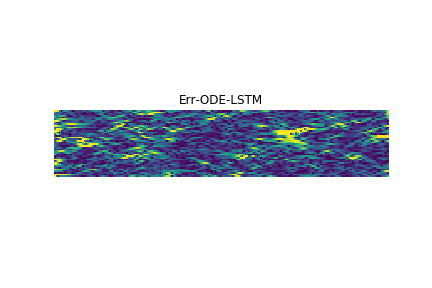}&
    \includegraphics[trim={55 110 150 110},clip, width=3cm]{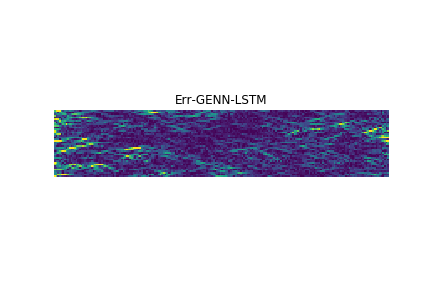}\\
 \end{tabular}
    \caption{{\bf Reconstruction of Lorenz-96 dynamics.} Upper left panel: solvers energy pathways using a predefined ODE-based cost or a learnable variational formulation. Upper right panel: associated pdfs of the reconstruction errors.  Lower panel: We depict the first half of an example of 200-time-step series of 40-dimensional Lorenz-96 states, the x-axis being the time axis; first row, reference states, FSGD-based reconstruction for variational cost (\ref{eq: var model}) with the true ODE model, reconstruction using the learned LSTM solver for this ODE-based setting, reconstruction when jointly learning the variational representation and the LSTM solver; second row, observation with missing data, absolute error maps of each of the 3 reconstructions. All states and errors are displayed with the same colormap.}
    \label{fig:res L96}
    \vspace*{-0.25cm}
\end{figure*}

\section{Related work}
\label{sec:rw}

{\bf Direct learning of inverse models:} Supervised learning has been increasingly used to address inverse problems by learning the relationship between observations and true states 
\cite{mccann2017convolutional,lucas2018using}. Interestingly, learnable data-driven solvers have been derived from standard model-driven algorithms, e.g. reaction-diffusion PDEs \cite{chen_learning_2015}, ADMM algorithm \cite{yang_deep_2016}. Such approaches do not however learn some underlying variational representation beyond the learned solver. 


{\bf Plug-and-play and learning-based regularizers:} In order to make learning approaches more versatile, and not specialized to a particular instance of a problem, so-called plug-and-play methods have been designed~\cite{venkatakrishnan2013plug}, including for instance denoising neural networks~\cite{meinhardt2017learning} and adversarial regularizers~\cite{lunz2018adversarial}. These regularizers trained offline can be incorporated to proximal optimization algorithms. 
This method can also lead to convergence guarantees of the resulting variational framework, provided the learned operator is sufficiently regular~\cite{ryu2019plug}. 

{\bf Bi-level optimization for inverse problems:} Recently, some learning schemes for inverse problems took a meta-learning~\cite{hospedales_meta-learning_2020} orientation, in the form of bi-level optimization problems. A given variational cost function (or a pre-trained regularizer such as with plug-and-play algorithms) forms an inner optimization problem, which is not solved directly as in classical variational schemes, but involves a learning-based solver, 
so that the reconstruction error is minimized in an outer loop~\cite{liu_bilevel_2019, adler2017solving}.
Note that a recent monograph covers most of the aspects discussed in this section so far \cite{arridge2019solving}.

{\bf Optimizer learning:} Learning solvers to minimize cost functions, in a so-called "learning-to-learn" fashion \cite{andrychowicz_learning_2016}, has attracted a lot of attention recently. The idea is to unroll an optimization algorithm (e.g., a descent direction algorithm), where the descent direction is parameterized by a RNN. The parameters of the RNN are learnt to to minimize the objective function. LSTMs are a popular choice since they allow to keep a memory of the past gradient information, as in momentum-based algorithms such as Adam. They also lead to sample-efficient algorithms and faster convergence than classical optimization algorithms \cite{andrychowicz_learning_2016}, as evidenced in our experiments. Optimizer learning is related to the bi-level optimization framework \cite{franceschi2018bilevel}, as the formal bi-level optimization setting can be relaxed by replacing the inner minimization through an unrolled optimization algorithm. Our learning strategy (\ref{eq: E2E loss}) can also be regarded as a relaxed version of a bi-level optimization problem
\begin{equation}
   \arg \min_{\Phi} \sum_n {\cal{L}} (x_n,\tilde{x}_n) \mbox{  s.t.  } 
   \tilde{x}_n = \arg \min_x  U_\Phi \left ( x,y_n , \Omega_n\right)
\end{equation}
where the inner minimization is replaced with a fixed number of iterations of a gradient-based solver. Here, the inner cost function also has trainable parameters. This relaxation, combined with automatic differentiation, allows our approach to jointly learn the parameters of the variational cost and of the associated solver, instead of resorting to a sequential minimization as in classical bi-level frameworks. If we fix the inner cost function,
we can incorporate any direct learning-based inversion scheme or plug-and-play approach to be fined tuned via meta-learning. The potential interest is two-fold: a lower complexity of the trainable operators and a greater interpretability of the learnt variational representations compared with a sole solver.

\section{Conclusion}
\label{sec:conlusion}

To the best of our knowledge, the problems of learning a variational cost (or more restrictively, a regularizer) from data for inverse problems, and learning a solver for a given variational cost, have always been treated sequentially so far. Our approach allows the joint data-driven discovery of a variational formulation of the inverse problem and the associated solver. Our experiments show that this is a better strategy than only learning the solver using a pre-defined or pre-trained variational cost, or even than directly inverting with the true generative model. 

This work opens new research avenues regarding the definition and resolution of variational formulations. A variety of energy-based formulations may be investigated beyond regularization terms $\|x-\Phi(x)\|^2$ as considered here, including physics-informed and theory-guided formulations as well as probabilistic energy-based representations.




\section*{Acknowledgements}
This work was supported by CNES (grant OSTST-MANATEE), Microsoft (AI EU Ocean awards) and ANR Projects Melody and OceaniX, and exploited HPC resources from GENCI-IDRIS (Grant 2020-101030).

\bibliographystyle{plain}  




\newpage
\appendix
\section{Supplementary Material}
\subsection{Lorenz-63 case-study}

Lorenz-63 dynamics involve a three-dimensional state governed by the following ODE:
\begin{equation}
\label{eq:lorenz-63}
\left \{\begin{array}{ccl}
\frac{dX_{t,1}}{dt} &=&\sigma \left (X_{t,2}-X_{t,2} \right ) \\
\frac{dX_{t,2}}{dt}&=&\rho X_{t,1}-X_{t,2}-X_{t,1}X_{t,3} \\
\frac{dX_{t,3}}{dt} &=&X_{t,1}X_{t,2}-\beta X_{t,3}
\end{array}\right.
\end{equation}
We consider the standard parameterization, $\sigma =10$, $\rho=28$ and  $\beta=8/3$, which leads to a strange attractor. We may refer the reader to \cite{lguensat_analog_2017} for additional illustrations.

In our experiments, we simulate time series with a 0.01 time step using a RK45 integration scheme \cite{dormand_family_1980}. More precisely, We first simulate a 20000-time-step sequence from an initial condition within the attractor. We use time steps 0 to 150000 to randomly sample 10000 200-time-step sequences for the training dataset, and time steps 15000 to 2000 to  randomly sample 2000 200-time-step sequences for the test dataset. Observation data are generated similarly to \cite{lguensat_analog_2017} with a sampling every 8 time steps and a Gaussian additive noise with a variance of 2. 

\subsection{Lorenz-96 case-study}

Lorenz-96 dynamics involve a 40-dimensional state with a periodic boundary condition  governed by the following ODE
\begin{equation}
\label{eq:lorenz-96}
\frac{dx_{t,i}}{dt} = \left (x_{t,i+1}-x_{t,i-2} \right ) x_{t,i-1} - x_{t,i} + F
\end{equation}
with $i$ the index from 1 to 40 and $F$ a scalar parameter set to 8. These dynamics involve chaotic wave-like patterns as illustrated in Fig.\ref{fig:res L96}

In our experiments, we proceed similarly to Lorenz-63 case-study. We first simulate a 10000-time-step sequence from an initial condition within the attractor. 

we simulate time series with a 0.05 time step using a RK45 integration scheme \cite{dormand_family_1980}. The training and test datasets involve time series with 200 time steps. We use time steps 0 to 7500 to randomly sample 2000 200-time-step sequences for the training dataset, and time steps 7500 to 10000 to randomly sample 256 200-time-step sequences for the test dataset. Observation data are generated similarly to \cite{lguensat_analog_2017} with a sampling every 4 time steps, for only 20 randomly selected components over the 40 components of the state, and a Gaussian additive noise with a variance of 2. 
\end{document}